\documentclass[12pt,a4paper]{article}

\usepackage[utf8]{inputenc}
\usepackage[T1]{fontenc}
\usepackage{lmodern}

\usepackage[top=2.5cm, bottom=2.5cm, left=2.5cm, right=2.5cm]{geometry}

\usepackage{amsmath, amssymb, amsthm}
\usepackage{bm}          

\usepackage{graphicx}
\usepackage{booktabs}
\usepackage{array}
\usepackage{tabularx}
\usepackage{multirow}
\usepackage{xcolor}
\usepackage{colortbl}

\usepackage[colorlinks=true, linkcolor=blue, citecolor=blue, urlcolor=blue]{hyperref}
\usepackage{url}

\usepackage{setspace}
\usepackage[labelfont=bf, font=small, justification=justified]{caption}

\usepackage{microtype}
\usepackage{enumitem}
\usepackage{float}
\usepackage{placeins}    
\usepackage{tikz}        

\definecolor{gogreen}{RGB}{46,139,87}
\definecolor{nogored}{RGB}{200,50,50}
\definecolor{lightgray}{gray}{0.92}
\definecolor{headerblue}{RGB}{44,62,80}
\definecolor{rowgray}{gray}{0.95}

\newcommand{\bx}{\bm{x}}
\newcommand{\by}{\bm{y}}
\newcommand{\bz}{\bm{z}}
\newcommand{\DKL}{D_{\mathrm{KL}}}

\newcommand{\lprior}{\lambda_{\mathrm{prior}}}
\newcommand{\lFIC}{\lambda_{y}}

\DeclareRobustCommand{\GO}{\texorpdfstring{\textcolor{gogreen}{\textbf{GO}}}{GO}}
\DeclareRobustCommand{\NOGO}{\texorpdfstring{\textcolor{nogored}{\textbf{NO-GO}}}{NO-GO}}

\title{\textbf{Retrodictive Forecasting: A Proof-of-Concept for Exploiting Temporal Asymmetry in Time Series Prediction}}

\author{%
Cédric Damour\\
\small ENERGY-Lab, Université de La Réunion, Saint-Denis, La Réunion, France\\
\small \href{mailto:cedric.damour@univ-reunion.fr}{\texttt{cedric.damour@univ-reunion.fr}}
}

\date{} 

\begin{document}

\maketitle
\vspace{-0.8em}
\vspace{0.5em}

\begin{abstract}
\begingroup
\small
\begin{singlespace}
We propose a \emph{retrodictive forecasting} paradigm for time series:
instead of predicting the future from the past, we identify the future
that best explains the observed present via inverse MAP optimization over
a Conditional Variational Autoencoder (CVAE)~\cite{kingma2014vae}. This
conditioning is a statistical modeling choice for Bayesian inversion; it
does not assert that future events cause past observations. The approach
is theoretically grounded in an information-theoretic arrow-of-time measure related to stochastic thermodynamics. The symmetrized Kullback--Leibler
divergence between forward and time-reversed trajectory ensembles
provides both the conceptual rationale and an operational GO/NO-GO
diagnostic for applicability. We implement the paradigm as MAP
inference over an inverse CVAE with a learned RealNVP normalizing-flow
prior and evaluate it on six time series cases: four synthetic processes
with controlled temporal asymmetry and two ERA5 reanalysis
datasets~\cite{hersbach2020} (North Sea wind speed and solar
irradiance). The work makes four contributions: (i)~a formal retrodictive inference
formulation; (ii)~an inverse CVAE architecture with flow prior and
Forward-Inverse Chaining warm-start; (iii)~a model-free irreversibility
diagnostic; and (iv)~a falsifiable validation protocol with
pre-specified predictions across controlled synthetic and real-world processes. All four pre-specified predictions are empirically supported: the
diagnostic correctly classifies all six cases; the learned
flow prior improves over an isotropic Gaussian baseline on GO cases; the inverse MAP yields no spurious advantage on
time-reversible dynamics; and on irreversible GO cases, it achieves
competitive or superior RMSE relative to forward baselines, with a
statistically significant $17.7\%$ reduction over a forward MLP on ERA5
solar irradiance (Diebold--Mariano test, $p < 0.001$). These results
provide a structured proof-of-concept that retrodictive
forecasting can constitute a viable alternative to conventional forward prediction when
statistical time-irreversibility is present and exploitable.

\end{singlespace}
\endgroup
\end{abstract}

\vspace{0.7em}

\section{Introduction}

Time series forecasting is conventionally framed as a forward problem:
given past observations $\bx_{t-n:t}$, predict future values
$\by_{t+1:t+m}$. This framing is so natural that its optimality is
rarely questioned~\cite{lim2021}. We ask whether the direction of
inference is optimal, or whether a fundamentally different
approach---identifying the future that would best explain the observed
present, in a purely statistical sense---could be advantageous in the
right conditions. This asymmetry between forward prediction and
retrodictive inference is not merely philosophical: it is measurable,
grounded in statistical thermodynamics, and potentially exploitable for
forecasting.

The central hypothesis of this work is that, for temporally irreversible
stochastic processes, a retrodictive approach---train an inverse
conditional generative model $p_\theta(\bx \mid \by, \bz)$ that
reconstructs the past from a candidate future, then at inference find
the future $\hat{\by}$ that maximises the retrodictive likelihood under
a learned prior $p_\psi(\by)$---is theoretically well-posed and
practically competitive. The key condition is statistical
irreversibility of the data-generating process, quantified by the
path-space Kullback--Leibler divergence between forward and
time-reversed trajectory ensembles (arrow-of-time diagnostic).

This paper makes four main contributions:

\begin{enumerate}[label=(\roman*)]
\item \textbf{Retrodictive forecasting paradigm.} We formalise time
  series forecasting as retrodictive inference: rather than learning a
  forward mapping from past to future, we learn an inverse conditional
  generative model $p_\theta(\bx \mid \by, \bz)$ that reconstructs the
  past from a candidate future, and we identify the forecast as the
  future $\hat{\by}$ that maximises the retrodictive likelihood under a
  learned prior $p_\psi(\by)$. This reframing shifts the computational
  burden from training to inference, and naturally accommodates
  multi-modal and discontinuous futures.

\item \textbf{Inverted CVAE with learned normalizing-flow prior.} We
  implement the paradigm as a Conditional Variational Autoencoder whose
  decoder is deliberately inverted---conditioned on the future to
  reconstruct the past---coupled with a RealNVP prior~\cite{dinh2017}
  over the future latent space. Forecasting is cast as a MAP
  optimization problem solved via Adam~\cite{kingma2015adam} with
  multi-start restarts, and optionally warm-started through
  Forward-Inverse Chaining (FIC).

\item \textbf{Arrow-of-time diagnostic and GO/NO-GO decision gate.}
  We propose a model-free irreversibility detector based on the
  symmetrized KL divergence (J-divergence) between forward and backward
  embedding distributions, inspired by stochastic
  thermodynamics~\cite{kawai2007,seifert2012}. This diagnostic serves
  as a pre-inference gate, predicting where the retrodictive paradigm is
  applicable and providing a falsifiable criterion for experimental
  validation.

\item \textbf{Controlled falsifiable validation suite.} We evaluate the
  paradigm against four pre-specified, falsifiable predictions on a
  controlled suite of four synthetic processes and two real-world ERA5
  meteorological datasets. We document both success cases and
  predictable failure modes, providing a replicable reference
  implementation.
\end{enumerate}

The remainder of the paper is organized as follows. Section~\ref{sec:formulation}
formalises the mathematical distinction between the classic and retrodictive
paradigms. Section~\ref{sec:theory} provides the theoretical justification via
the arrow-of-time and thermodynamic fluctuation theorems.
Section~\ref{sec:architecture} details the CVAE architecture and MAP
inference algorithm. Section~\ref{sec:related} reviews related work.
Section~\ref{sec:diagnostic} introduces the arrow-of-time diagnostic and
GO/NO-GO gate. Section~\ref{sec:suite} describes the proof-of-concept
process suite. Section~\ref{sec:protocol} presents the evaluation protocol
and falsifiable predictions. Section~\ref{sec:results} reports all results.
Section~\ref{sec:discussion} discusses implications, limitations, and
architectural extensions. Section~\ref{sec:conclusion} concludes. This work is intentionally positioned as a proof-of-concept: success is defined as demonstrating the empirical viability and structural coherence of the retrodictive paradigm under measurable time-irreversibility conditions, rather than outperforming state-of-the-art forecasters on public leaderboards.

\section{Mathematical Formulation of Both Paradigms}
\label{sec:formulation}

Let $\{s_t\}_{t=1}^{T}$ be a univariate time series. We define the
supervised pair $(\bx_t, \by_t)$ using a sliding window of past length
$n = 32$ and forecast horizon $m = 16$:
\begin{align}
  \bx_t &= [s_t, \ldots, s_{t+n-1}] \in \mathbb{R}^n, \\
  \by_t &= [s_{t+n}, \ldots, s_{t+n+m-1}] \in \mathbb{R}^m.
\end{align}

\subsection{Classic paradigm (forward prediction)}

A forward model $f_\theta : \mathbb{R}^n \to \mathbb{R}^m$ is trained
by minimizing MSE on training pairs. Inference is a single forward pass:
$\hat{\by} = f_\theta(\bx_{\mathrm{obs}})$. Inference time is $O(1)$.

\subsection{Retrodictive paradigm (inverse inference)}

The training phase is reversed: a conditional generative model
$p_\theta(\bx \mid \by, \bz)$ is trained to reconstruct the past from
the future and a latent variable $\bz$. At inference, we solve:
\begin{equation}
  (\hat{\by}, \hat{\bz}) = \arg\min_{\by, \bz}
    \Bigl[
      -\log p_\theta(\bx_{\mathrm{obs}} \mid \by, \bz)
      - \lambda \log p(\by)
      - \log p(\bz)
    \Bigr],
  \label{eq:map}
\end{equation}
where $p(\by)$ is a learned RealNVP flow prior on futures and $p(\bz) =
\mathcal{N}(\mathbf{0}, \mathbf{I})$. (Here $\lambda$ corresponds to
$\lprior$ in Section~\ref{sec:architecture}; the FIC term $\lFIC$ is
omitted for clarity.) Inference time is $O(K \times N_{\mathrm{MAP}})$
where $K = 5$ restarts and $N_{\mathrm{MAP}} = 200$ steps.

\paragraph{Causality clarification.}
Conditioning on $\by_{\mathrm{future}}$ in the generative model
$p_\theta(\bx \mid \by, \bz)$ is a modeling choice for statistical
inverse inference, not an assertion that future events physically cause
past observations. During training, both $\bx_{\mathrm{past}}$ and
$\by_{\mathrm{future}}$ are extracted from historical sequences in which
both quantities are already observed; no information from the true
future relative to deployment time is used. During inference,
$\by_{\mathrm{future}}$ is not observed but treated as a free
optimization variable---a hypothesis about what the future might
be---whose plausibility is controlled by the learned prior $p_\psi(\by)$
and whose consistency with the observed past is measured by the
retrodictive likelihood. The method thus operates entirely within the
observational regime of Pearl's causal hierarchy~\cite{pearl2009}: it
performs Bayesian inversion over a learned statistical model, not causal
intervention. The term \emph{retrodictive} is used deliberately to
distinguish this inference from both forward prediction and causal
reasoning.

\subsection{Conceptual distinction and relationship to 4D-Var}

The retrodictive paradigm differs fundamentally from 4D-Var data
assimilation~\cite{talagrand1987}: 4D-Var estimates an initial state
consistent with a known physical dynamical model and observations,
whereas retrodictive forecasting estimates a future state consistent
with a learned data-driven generative model and an observed past. The
forward operator is entirely agnostic to explicit physical equations,
making the approach applicable to systems where governing dynamics are
unknown or intractable.

\section{Theoretical Justification: The Arrow of Time}
\label{sec:theory}

\subsection{Statistical irreversibility and the forecasting advantage}

For a stochastic process $\{s_t\}$, define the path-space J-divergence
(symmetrized KL) at block window length $w$:
\begin{equation}
  J(w) = \tfrac{1}{2}\DKL(p_{\mathrm{fwd}} \| p_{\mathrm{bwd}})
        + \tfrac{1}{2}\DKL(p_{\mathrm{bwd}} \| p_{\mathrm{fwd}}),
  \label{eq:jdiv}
\end{equation}
where $p_{\mathrm{fwd}}$ and $p_{\mathrm{bwd}}$ are the empirical
distributions of forward and time-reversed $2w$-length windows,
respectively, estimated from either the raw series (LEVEL representation)
or its first differences (DIFF representation). A large $J(w) > 0$
indicates temporal irreversibility: knowing the present state constrains
the admissible past trajectories more strongly than it determines future
evolutions. This provides a necessary structural condition under which
retrodictive advantage may arise.

\subsection{Thermodynamic grounding}

Under Markov dynamics and local detailed balance, $J(w)$ equals the
mean entropy production rate per
window~\cite{kawai2007,seifert2012,crooks1999,parrondo2009}. In
practice, with coarse-grained observations, $J(w)$ provides a lower
bound on thermodynamic irreversibility and is used as an operational
statistical diagnostic. We emphasize that the identification between
J-divergence and entropy production is exact only under stochastic
thermodynamics assumptions. In coarse-grained real-world geophysical
time series such as ERA5, these assumptions are not strictly satisfied.
Accordingly, $J(w)$ should be interpreted as an information-theoretic
measure of statistical time-irreversibility rather than a direct
physical entropy production estimator.

\subsection{Conditions for applicability --- GO/NO-GO diagnostic}

The retrodictive paradigm is expected to provide a meaningful advantage
only when (i)~the process is statistically irreversible ($J(w) > 0$ at
the relevant scales), and (ii)~the forecast horizon is within the
decorrelation time of the process. The arrow-of-time diagnostic
(Section~\ref{sec:diagnostic}) provides an operational GO/NO-GO gate
based on block permutation testing of $J(w)$ at multiple scales and in
two representations (LEVEL and DIFF), enabling a model-free,
pre-inference classification of each process.

\section{Architecture: Inverted Conditional Variational Autoencoder}
\label{sec:architecture}

\subsection{Design rationale}

A standard CVAE models $p(\by \mid \bx)$---the conditional distribution
of futures given the past. Our inverse CVAE reverses the conditioning:
it models $p_\theta(\bx \mid \by, \bz)$, enabling the decoder to
reconstruct the past from the future and a latent variable $\bz$. This
conditional decoder formulation is critical: it ensures that MAP
inference over $(\by, \bz)$ optimizes the retrodictive likelihood while
maintaining a coherent generative structure.

\subsection{Architecture components}

\textbf{Posterior encoder $q_\phi(\bz \mid \bx, \by)$:} a 2-layer MLP
with hidden size 128 that maps (past, future) pairs to the posterior
parameters $(\bm{\mu}, \bm{\sigma})$ of the latent variable $\bz \in
\mathbb{R}^8$.

\textbf{RealNVP flow prior $p_\psi(\by)$:} a RealNVP normalizing flow
(8 coupling layers, alternating masking) that models the marginal prior
over futures $\by$~\cite{dinh2017}. At inference, this flow prior
$p_\psi(\by)$ serves as the regularizer term $-\log p(\by)$ in the MAP
objective~\eqref{eq:map}. This design enables the prior to capture the
non-Gaussian, structured geometry of the marginal distribution of
futures, which is critical for effective MAP optimization.

\textbf{Conditional decoder $p_\theta(\bx \mid \by, \bz)$:} a 2-layer
MLP with hidden size 128 that outputs Gaussian parameters $(\bm{\mu},
\bm{\sigma})$ for the past reconstruction given $(\by, \bz)$. The
conditional dependency on $\by$ is the key architectural choice that
couples the decoder to the future being optimized during MAP inference.

\subsection{Training}

The inverse CVAE is trained to maximize the ELBO:
\begin{equation}
  \mathcal{L}(\theta, \phi)
  = \mathbb{E}\bigl[\log p_\theta(\bx \mid \by, \bz)\bigr]
    - \beta\, \mathrm{KL}\bigl(q_\phi(\bz \mid \bx, \by) \| \mathcal{N}(\mathbf{0},\mathbf{I})\bigr),
  \label{eq:elbo}
\end{equation}
with $\beta = 1$ and Adam optimizer ($\mathrm{lr} = 2 \times 10^{-3}$,
80~epochs). The latent prior $p(\bz) = \mathcal{N}(\mathbf{0},\mathbf{I})$
is standard Gaussian; no conditional prior network $p_\psi(\bz \mid \by)$
is learned.

\subsection{MAP inference and Forward-Inverse Chaining}
\label{sec:fic}

At inference, we solve:
\begin{equation}
  (\hat{\by}, \hat{\bz}) = \arg\min_{\by, \bz}
    \Bigl[
      -\log p_\theta(\bx_{\mathrm{obs}} \mid \by, \bz)
      + \lprior \bigl(-\log p_\psi(\by)\bigr)
      + \lFIC \|\by - \hat{\by}_{\mathrm{FIC}}\|^2
    \Bigr].
  \label{eq:map_full}
\end{equation}
The Adam optimizer ($\mathrm{lr} = 5 \times 10^{-2}$, gradient clipping
norm~$5.0$) runs for $N = 200$ steps with $K = 5$ random restarts and $\lprior = 2.0$. The
Forward-Inverse Chaining (FIC) innovation initializes $\hat{\by}_{\mathrm{FIC}}$
from the forward CVAE prediction, providing a warm-start that
accelerates convergence and improves optimization stability. The FIC warm-start
operates via initialisation only: restart $k=0$ is initialised from the
forward CVAE prediction $\hat{\by}_{\mathrm{FIC}}$, while restarts
$k=1\ldots K-1$ use random samples from the prior. 

\section{Related Work}
\label{sec:related}

The retrodictive forecasting paradigm sits at the intersection of three
established research streams.

\paragraph{Inverse problems and deep learning.}
Solving inverse problems---recovering latent variables from observed
data---is a classical challenge in geophysics, imaging, and
engineering~\cite{ongie2020}. Recent advances in invertible neural
networks for Bayesian geophysical inversion~\cite{zhang2021} and in
normalizing-flow-based inverse modeling~\cite{ardizzone2019} demonstrate
how learned generative models can replace explicit physical forward
operators. Our contribution differs: rather than inverting a known
physical model, we formulate forecasting itself as a statistical inverse
problem.

\paragraph{Variational autoencoders for time series.}
CVAEs have been widely used for probabilistic time series forecasting,
including wind energy prediction~\cite{salazar2022} and wind turbine
fatigue estimation~\cite{mylonas2021}. These approaches model $p(\by
\mid \bx)$ and generate futures conditioned on the past. In contrast,
our framework reverses the conditioning direction and learns
$p_\theta(\bx \mid \by, \bz)$, changing the inference semantics
fundamentally.

\paragraph{Statistical time-reversal asymmetry.}
Time-reversal asymmetry in time series has traditionally been studied as
a diagnostic for nonlinearity or dynamical
irreversibility~\cite{theiler1992}. Our work adopts the symmetrized KL
divergence (J-divergence) between forward and backward embedding
distributions as an operational irreversibility measure. Statistical
significance is assessed using a block permutation test grounded in
surrogate-data methodology~\cite{schreiber2000}, yielding a model-free
GO/NO-GO diagnostic.

\section{Arrow-of-Time GO/NO-GO Diagnostic (Irreversibility Gate)}
\label{sec:diagnostic}

\subsection{Motivation and role}

The arrow-of-time diagnostic quantifies statistical
time-irreversibility in a univariate time series. It serves as a
pre-inference gate with three purposes: (i)~provide a reproducible,
model-free criterion for separating GO processes (strong directional
structure) from NO-GO ones (time-symmetric); (ii)~establish a
falsifiable failure expectation---in NO-GO regimes, inverse inference
should be ill-posed with no structural advantage; (iii)~offer an
interpretable complement to retrodictive performance metrics.

\subsection{Construction: forward and backward embeddings}

Given a univariate series $\{s_t\}$ and half-window length $w$, we
construct at each admissible index $t$ a pair of embedding vectors in
$\mathbb{R}^{2w}$:
\begin{align}
  \mathbf{e}^{(f)}_t &= [s_t, s_{t+1}, \ldots, s_{t+w-1},\; s_{t+w}, \ldots, s_{t+2w-1}], \\
  \mathbf{e}^{(b)}_t &= [s_{t+2w-1}, \ldots, s_{t+w},\; s_{t+w-1}, \ldots, s_t],
\end{align}
where $\mathbf{e}^{(f)}_t$ concatenates a past segment with a future
segment, and $\mathbf{e}^{(b)}_t$ is its time-reversal. Two
representations are used in parallel: LEVEL (raw series $s_t$) and DIFF
(first differences $\Delta s_t = s_{t+1} - s_t$).

\subsection{J-divergence and block permutation test}

The divergence between the forward and backward point clouds $F_w$ and
$B_w$ is quantified using the symmetrized KL divergence
(Eq.~\eqref{eq:jdiv}), estimated via $k$-nearest-neighbor density
estimation~\cite{perezcruz2008,kraskov2004}. Statistical significance is
assessed by a block permutation test: the null hypothesis of
time-reversibility is that $F_w$ and $B_w$ are exchangeable. Under
permutation of blocks (block length $= w$), the J-divergence under the
null is estimated by Monte Carlo (500 permutations), yielding a p-value
$p_{\mathrm{perm}}$.

A test is declared significant at scale (representation, window $w$) if
$p_{\mathrm{perm}} < \alpha = 0.05$. The overall verdict is GO if at
least $C_{\min} = 2$ out of 3 window scales ($w \in \{2, 4, 8\}$) are
significant for at least one representation (LEVEL or DIFF). Otherwise,
the verdict is NO-GO.

We define the scalar summary $\Delta_{\mathrm{arrow}}$ as the median of $J_{\mathrm{obs}}(w)$ 
across all tested scales $w \in \{2, 4, 8\}$ and both representations (LEVEL and DIFF):
\begin{equation}
    \Delta_{\mathrm{arrow}} = \mathrm{median}_{w,\, \mathrm{rep}} \; J_{\mathrm{obs}}(w, \mathrm{rep}).
    \label{eq:delta_arrow}
\end{equation}
This scalar provides a single-number summary of irreversibility strength, 
used for cross-case comparison (Fig.~\ref{fig:06}). The GO/NO-GO verdict is determined 
by the significance criterion above, not by a threshold on $\Delta_{\mathrm{arrow}}$ directly.

\subsection{Multi-scale and multi-representation design}

Three window scales ($w = 2, 4, 8$) and two representations (LEVEL,
DIFF) give 6 independent test instances per case. This multi-scale
design guards against scale-specific artefacts while providing
sensitivity to a broad range of irreversibility mechanisms. The LEVEL
representation captures amplitude-level asymmetries; DIFF captures
transition-level asymmetries. Requiring consensus across $C_{\min} = 2$
scales within one representation reduces the false-positive rate while
maintaining power for genuine irreversibility.

\subsection{Theoretical properties}

The J-divergence is: (i)~reparametrization-invariant; (ii)~
positive when the process is statistically irreversible; (iii)~zero if
and only if forward and backward distributions are identical; and
(iv)~related to thermodynamic entropy production under stochastic
thermodynamics assumptions~\cite{kawai2007,seifert2012}.

\section{Synthetic and Real-World Processes (Proof-of-Concept Suite)}
\label{sec:suite}

\subsection{Design rationale}

A controlled experimental environment is essential for establishing the empirical viability and operational falsifiability of the retrodictive paradigm. We evaluate the framework on four synthetic time series with fully
specified generative mechanisms and known temporal asymmetry properties,
enabling sharp falsifiable predictions. Two ERA5 real-world cases extend
validity to geophysical observations. The suite satisfies four design
principles: (P1)~coverage of distinct irreversibility mechanisms;
(P2)~falsifiability via known NO-GO cases with $\Delta_{\mathrm{arrow}} \approx 0$;
(P3)~cross-disciplinary transferability; and (P4)~symmetric Gaussian
innovations in all synthetic cases so that irreversibility arises
exclusively from dynamics.

\subsection{Synthetic cases}

\paragraph{Case A --- \GO{} dissipative NLAR.}
A nonlinear autoregressive process with state-dependent multiplicative noise:
\begin{equation}
  s_t = \alpha\,\tanh(s_{t-1}) + \gamma_q\, s_{t-1}^2 + \gamma_c\, s_{t-1}^3
        + \sigma(s_{t-1})\,\varepsilon_t,
  \label{eq:caseA}
\end{equation}
where $\sigma(s_{t-1}) = \sigma_0 + \sigma_1|s_{t-1}|$ and
$\varepsilon_t \sim \mathcal{N}(0,1)$. Parameters: $\alpha = 0.7$,
$\gamma_q = 0.05$, $\gamma_c = -0.08$, $\sigma_0 = 0.3$, $\sigma_1 =
0.35$. Three mechanisms jointly produce time-reversal asymmetry: the
$\tanh$ contraction, the cubic dissipation term, and state-dependent
multiplicative noise. This case models driven-dissipative physical
systems such as atmospheric boundary layer turbulence.

\paragraph{Case B --- \NOGO{} symmetric random walk.}
$s_t = s_{t-1} + \sigma\,\varepsilon_t$, $\varepsilon_t \sim
\mathcal{N}(0,1)$, $\sigma = 0.50$. This process is exactly
time-reversible by construction.

\paragraph{Case C --- \GO{} shot-noise relaxation.}
An excitation-relaxation process with rare Poisson-driven jumps:
\begin{align}
  x_t &= \lambda\, x_{t-1} + A_t, \\
  s_t &= x_t + \sigma_{\mathrm{obs}}\,\varepsilon_t,
\end{align}
where $\lambda = 0.95$, $A_t = J_t$ with probability $p = 0.04$ (else
$0$), $J_t \sim \mathrm{Exp}(\mu_J = 1.5)$, and $\sigma_{\mathrm{obs}}
= 0.05$. Designed to model cloud-induced irradiance variations.

\paragraph{Case D --- \NOGO{} noisy sinusoid.}
$s_t = A\sin(2\pi t / P) + \sigma\,\varepsilon_t$, $\varepsilon_t \sim
\mathcal{N}(0,1)$. Parameters: $A = 1$, $P = 40$~steps, $\sigma = 0.5$.
Time-reversible at all scales.

\subsection{Real-world ERA5 cases}

\paragraph{Case ERA5 --- North Sea 10\,m wind speed.}
Hourly wind speed at 10\,m from ERA5 reanalysis, location $56^\circ$N
$3^\circ$E (North Sea), year 2023. $T = 8{,}736$ hourly observations.
Wind speed is log-transformed and standardized. \GO{} via the DIFF
representation ($J_{\mathrm{obs}} = 0.55$).

\paragraph{Case ERA\_ssrd --- North Sea solar irradiance.}
Surface Solar Radiation Downwards (SSRD) from ERA5, $56^\circ$N
$5^\circ$E, year 2023, converted to W/m$^2$. A daylight filter is
applied (solar zenith angle $< 80^\circ$), retaining $T \approx 5{,}110$
daytime hourly observations. Strongly \GO{} via DIFF ($J_{\mathrm{obs}}
= 5.30$). The complete process suite is summarised in
Table~\ref{tab:suite}.

\begin{table}[!t]
\centering
\caption{Process suite summary. All six cases with generative mechanism,
noise structure, expected irreversibility, and key parameters.}
\label{tab:suite}
\footnotesize\setlength{\tabcolsep}{3pt}
\renewcommand{\arraystretch}{1.2}
\begin{tabularx}{\linewidth}{p{1.5cm} p{1.1cm} p{2.0cm} p{2.0cm} >{\raggedright\arraybackslash}X p{2.6cm} >{\raggedright\arraybackslash}X}
\rowcolor{headerblue}
\textcolor{white}{\textbf{Case}} &
\textcolor{white}{\textbf{Verdict}} &
\textcolor{white}{\textbf{Process}} &
\textcolor{white}{\textbf{Noise structure}} &
\textcolor{white}{\textbf{Irreversibility mechanism}} &
\textcolor{white}{\textbf{Irreversibility strength ($\Delta_{\mathrm{arrow}}$)}} &
\textcolor{white}{\textbf{Key parameters}}\\
\rowcolor{white}
\textbf{A} & \GO{} & NLAR dissipative & \parbox[t]{\linewidth}{$\sigma(s){=}\sigma_0{+}$\\$\sigma_1|s|$} &
  tanh contraction + cubic dissipation + multiplicative noise & Strong (LEVEL) &
  $\alpha{=}0.7$, $\gamma_q{=}0.05$, $\gamma_c{=}{-}0.08$, $\sigma_0{=}0.3$, $\sigma_1{=}0.35$\\
\rowcolor{rowgray}
\textbf{B} & \NOGO{} & Symmetric RW & Gaussian increments & None & ${\approx}\,0$ &
  $T{=}20$k, $\sigma{=}0.50$\\
\rowcolor{white}
\textbf{C} & \GO{} & Shot-noise relaxation & Excitation $p{=}0.04$, exp.\ relax. &
  Causal excitation-relax. & Very strong (LEVEL+DIFF) &
  $T{=}20$k, decay$=0.95$, $p_{\mathrm{shot}}{=}0.04$, shot\_scale$=1.5$, $\sigma_{\mathrm{obs}}{=}0.05$\\
\rowcolor{rowgray}
\textbf{D} & \NOGO{} & Noisy sinusoid & Symmetric noise on period & None (symmetric) &
  ${\approx}\,0$ & $T{=}20$k, $A{=}1.0$, $P{=}40$, $\sigma{=}0.50$\\
\rowcolor{white}
\textbf{ERA5} & \GO{} & North Sea wind (10m) & ERA5 reanalysis 2023 &
  Turbulence cascade & Moderate (DIFF) &
  $T{=}8{,}736$h, $56^\circ$N $3^\circ$E\\
\rowcolor{rowgray}
\textbf{ERA\_ssrd} & \GO{} & North Sea solar (SSRD) & ERA5 reanalysis 2023 &
  Cloud attenuation & Strong (DIFF) &
  $T{\approx}5{,}110$h (daylight), $56^\circ$N $5^\circ$E\\
\end{tabularx}
\renewcommand{\arraystretch}{1.0}
\end{table}

\FloatBarrier

\section{Evaluation Protocol}
\label{sec:protocol}

\subsection{Data pipeline and experimental setup}

For each synthetic process (Cases A--D), a single univariate series
$\{s_t\}$ of length $T = 20{,}000$ is generated with a fixed random
seed (seed~$= 42$). For the two real-world cases, the series are
extracted from ERA5 reanalysis data: Case ERA5 uses 10-metre wind speed
(North Sea, $56^\circ$N $3^\circ$E, 2023; $T = 8{,}736$~h), and Case
ERA\_ssrd uses surface solar radiation downwards converted to W/m$^2$
with a daylight filter applied (North Sea, $56^\circ$N $5^\circ$E,
2023; $T \approx 5{,}110$~h). Supervised pairs are extracted using a
sliding window with past window $n = 32$ and forecast horizon $m = 16$,
split chronologically and without shuffling into training (70\%),
validation (15\%), and test (15\%). All windows are standardized using
z-score normalization computed on the training set only.

\subsection{Methods compared}

Five methods are evaluated on each case, enabling systematic ablation
of the retrodictive paradigm's components (Table~\ref{tab:methods}).

\begin{table}[htbp]
\centering
\caption{Methods compared. Five evaluation methods enabling ablation of
retrodictive paradigm components.}
\label{tab:methods}
\small\setlength{\tabcolsep}{4pt}
\renewcommand{\arraystretch}{1.5}
\begin{tabularx}{\textwidth}{c >{\raggedright\arraybackslash}p{2.6cm} l l X}
\rowcolor{headerblue}
\textcolor{white}{\textbf{ID}} &
\textcolor{white}{\textbf{Name}} &
\textcolor{white}{\textbf{Training obj.}} &
\textcolor{white}{\textbf{Inference}} &
\textcolor{white}{\textbf{Description}}\\
\rowcolor{white}
\textbf{M1} & Na\"ive mean & --- & $\hat{\by} = \bar{\by}_{\mathrm{train}}$ &
  Unconditional mean of training futures. Zero-information baseline.\\
\rowcolor{rowgray}
\textbf{M2} & Forward MLP & MSE & $\hat{\by} = f(\bx)$ &
  2-hidden-layer MLP (128 units, ReLU), trained on $(\bx, \by)$ pairs.\\
\rowcolor{white}
\textbf{M3} & Forward CVAE & ELBO (forward) & $\hat{\by} \sim p_\theta(\by \mid \bx)$ &
  Standard CVAE with prior $p(\bz \mid \bx)$, decoder
  $p_\theta(\by \mid \bx, \bz)$. Samples averaged.\\
\rowcolor{rowgray}
\textbf{M4} & Inverse MAP (flow prior) & ELBO (inverse) &
  $(\hat{\by}, \hat{\bz}) = \mathrm{MAP}$ &
  Inverted CVAE with RealNVP flow prior. MAP over $(\by, \bz)$. \textit{Primary method.}\\
\rowcolor{white}
\textbf{M5} & Inverse MAP (N(0,I) prior) & ELBO (inverse) &
  $(\hat{\by}, \hat{\bz}) = \mathrm{MAP}$ &
  Same as M4 but with isotropic Gaussian prior $p(\by) = \mathcal{N}(\mathbf{0},\mathbf{I})$.
  Ablation for P2.\\
\end{tabularx}
\renewcommand{\arraystretch}{1.0}
\end{table}

\subsection{Evaluation metrics}

The primary comparison metric is the global RMSE in standardized space:
\begin{equation}
  \mathrm{RMSE} = \sqrt{\frac{1}{Nm}\sum_{i,j}(\hat{y}_{i,j} - y^{\mathrm{true}}_{i,j})^2}.
\end{equation}
All metrics are computed on held-out test data following established
out-of-sample evaluation practice~\cite{tashman2000,gneiting2007}.
Statistical significance is assessed via the Diebold--Mariano
test~\cite{diebold1995,harvey1997} (two-sided, Harvey et al.\ correction)
comparing each method against the forward MLP (M2). Retrodictive-specific
metrics include the per-sample RetroNLL
($-\log p_\theta(\bx_{\mathrm{obs}} \mid \hat{\by}, \hat{\bz})$) and
multi-start dispersion $\mathrm{std}(\hat{\by})$ across $K = 5$ restarts.

\subsection{Falsifiable predictions (pre-specified)}

Four falsifiable predictions are stated a priori before any experiment
is run. The tolerance band of $\pm 5\%$ on the RMSE ratio (Predictions
P3 and P4) accommodates finite-sample variance while maintaining
discriminative power. The four predictions are listed in
Table~\ref{tab:predictions}.

\begin{table}[htbp]
\centering
\caption{Falsifiable predictions. Four pre-specified predictions with
pass criteria and observed outcomes (Section~\ref{sec:results}).}
\label{tab:predictions}
\small\setlength{\tabcolsep}{4pt}
\renewcommand{\arraystretch}{1.5}
\begin{tabular}{p{0.6cm} p{2.9cm} p{1.8cm} p{1.8cm} p{3.3cm} p{3.4cm}}
\rowcolor{headerblue}
\textcolor{white}{\textbf{ID}} &
\textcolor{white}{\textbf{Statement}} &
\textcolor{white}{\textbf{Cases}} &
\textcolor{white}{\textbf{Metric}} &
\textcolor{white}{\textbf{Pass criterion}} &
\textcolor{white}{\textbf{Observed}}\\
\rowcolor{white}
\textbf{P1} & Arrow-of-time diagnostic matches expected verdicts &
  A--D, ERA5, ERA\_ssrd & \GO/\NOGO{} verdict &
  \GO: A,C,ERA5, ERA\_ssrd; \newline \NOGO: B,D &
  A,C,ERA5,ERA\_ssrd $\to$ \GO{}; \newline B,D $\to$ \NOGO\\
\rowcolor{rowgray}
\textbf{P2} & Flow prior improves inverse inference vs $\mathcal{N}(0,I)$ on \GO{} cases &
  A, C & RMSE &
  RMSE$_{\mathrm{flow}}<$RMSE$_{\mathcal{N}(0,I)}$ &
  A: $1.038 < 1.074$;\newline C: $0.782 < 0.872$\\
\rowcolor{white}
\textbf{P3} & No meaningful directional advantage on \NOGO{} cases &
  B, D & Ratio Inv/MLP & $\geq 0.95$ ($\pm 5\%$) &
  B: 1.870;\newline D: 0.984\\
\rowcolor{rowgray}
\textbf{P4} & Inverse MAP competitive with or beats MLP on \GO{} cases &
  A, C, ERA5, ERA\_ssrd & Ratio Inv/MLP & $\leq 1.05$ ($\pm 5\%$) &
  A: 0.897; C: 1.014;\newline ERA5: 0.991;\newline ERA\_ssrd: 0.823\\
\end{tabular}
\renewcommand{\arraystretch}{1.0}
\end{table}

\section{Results}
\label{sec:results}

\subsection{Overview}

\begin{figure}[htbp]
  \centering
  \includegraphics[width=\textwidth]{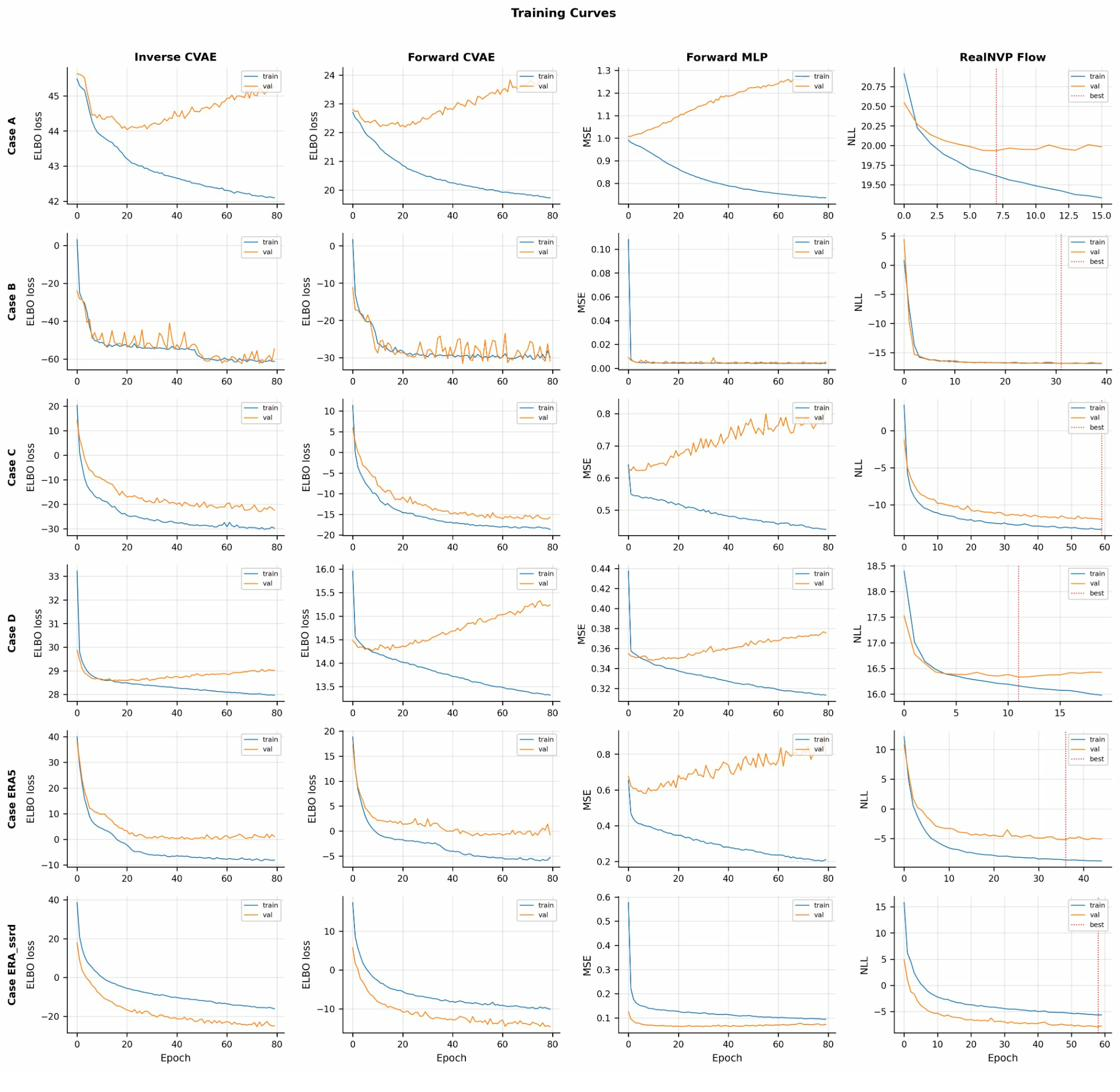}
  \caption{Training curves (ELBO loss) for all six cases. Solid:
  training loss; dashed: validation loss. Stable convergence without
  overfitting across all cases confirms correct CVAE training.}
  \label{fig:01}
\end{figure}

\begin{figure}[htbp]
  \centering
  \includegraphics[width=\textwidth]{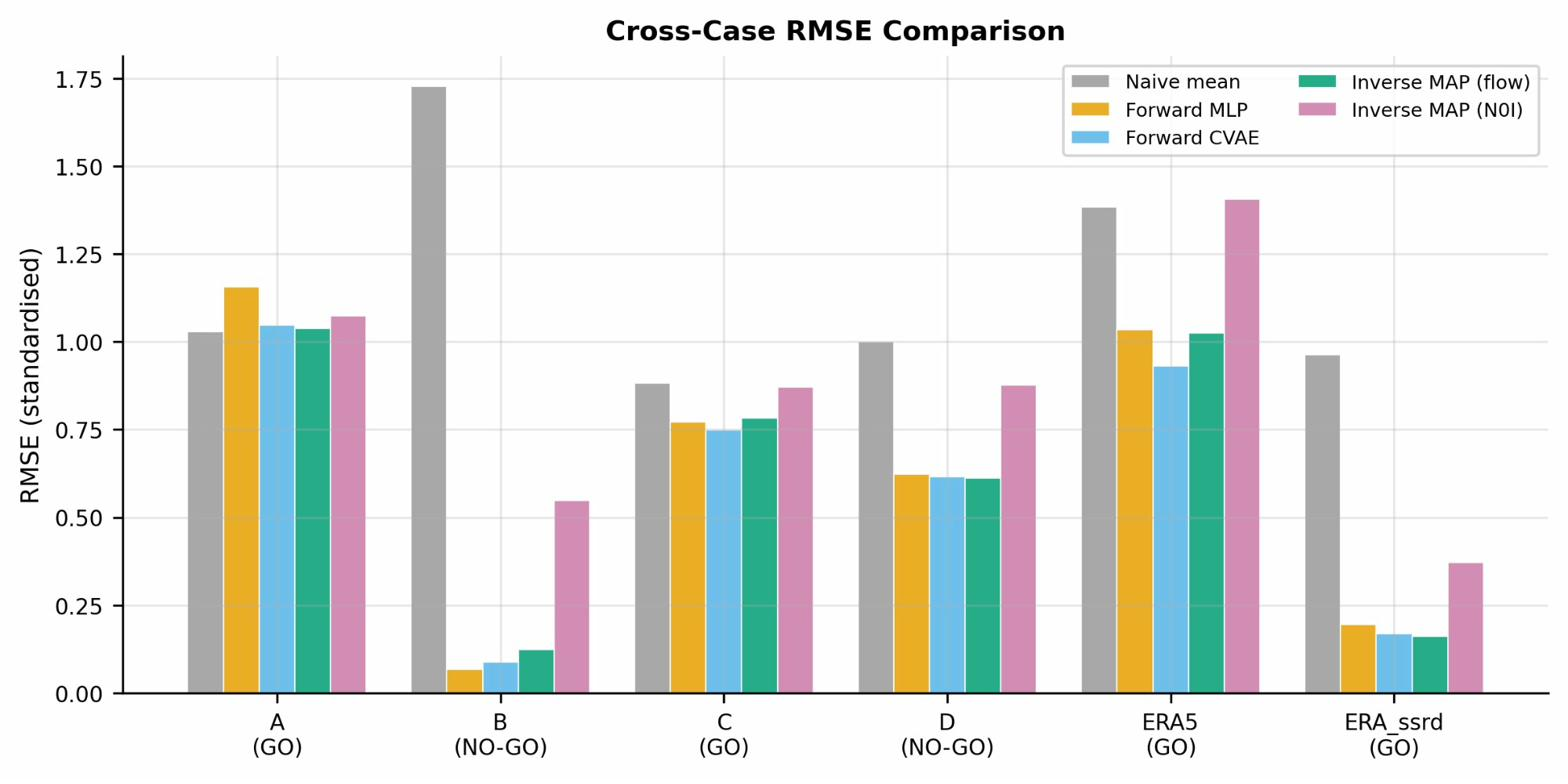}
  \caption{Cross-case RMSE comparison (all six cases, five methods).
  The inverse MAP (flow, green) is competitive or best on all \GO{}
  cases, while the forward MLP (orange) clearly dominates on Case~B
  (\NOGO). Case~D shows near-perfect equivalence across methods.}
  \label{fig:02}
\end{figure}

\begin{figure}[htbp]
  \centering
  \includegraphics[width=0.95\textwidth]{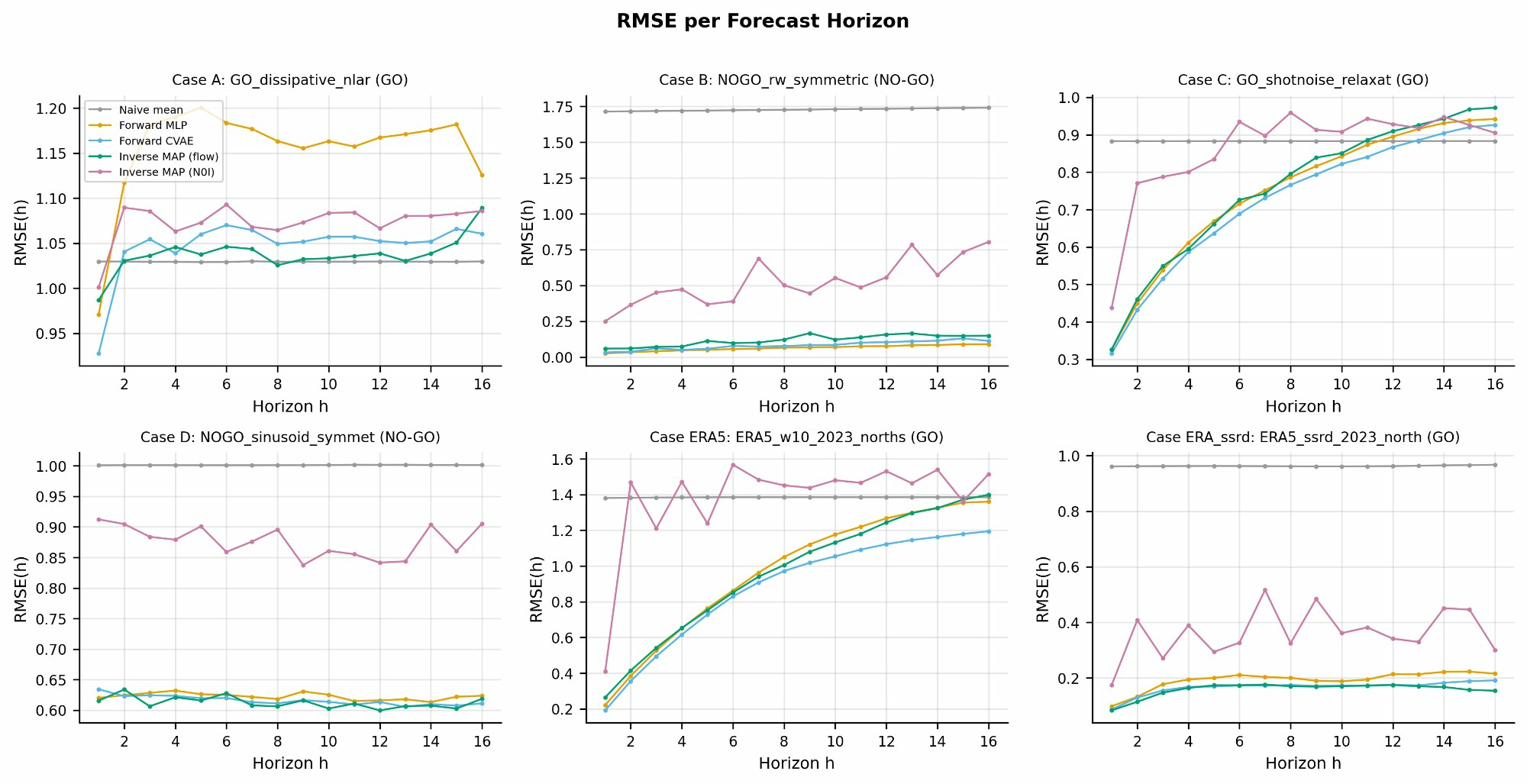}
  \caption{RMSE per forecast horizon $h \in \{1, \ldots, 16\}$ for all
  six cases and five methods. \GO{} cases (A, C, ERA5, ERA\_ssrd): the
  inverse MAP with flow prior (dark green) is competitive with or below
  the forward MLP (orange) across most horizons, with the advantage
  widening at longer horizons on ERA\_ssrd. \NOGO{} cases (B, D): the
  forward MLP dominates on Case~B; Case~D shows near-equivalence
  consistent with time-reversibility. The N(0,I) prior (pink) uniformly
  underperforms the flow prior on \GO{} cases. Compare with
  Fig.~\ref{fig:10} (\GO{} cases only, overlay).}
  \label{fig:03}
\end{figure}

\begin{figure}[htbp]
  \centering
  \includegraphics[width=0.75\textwidth]{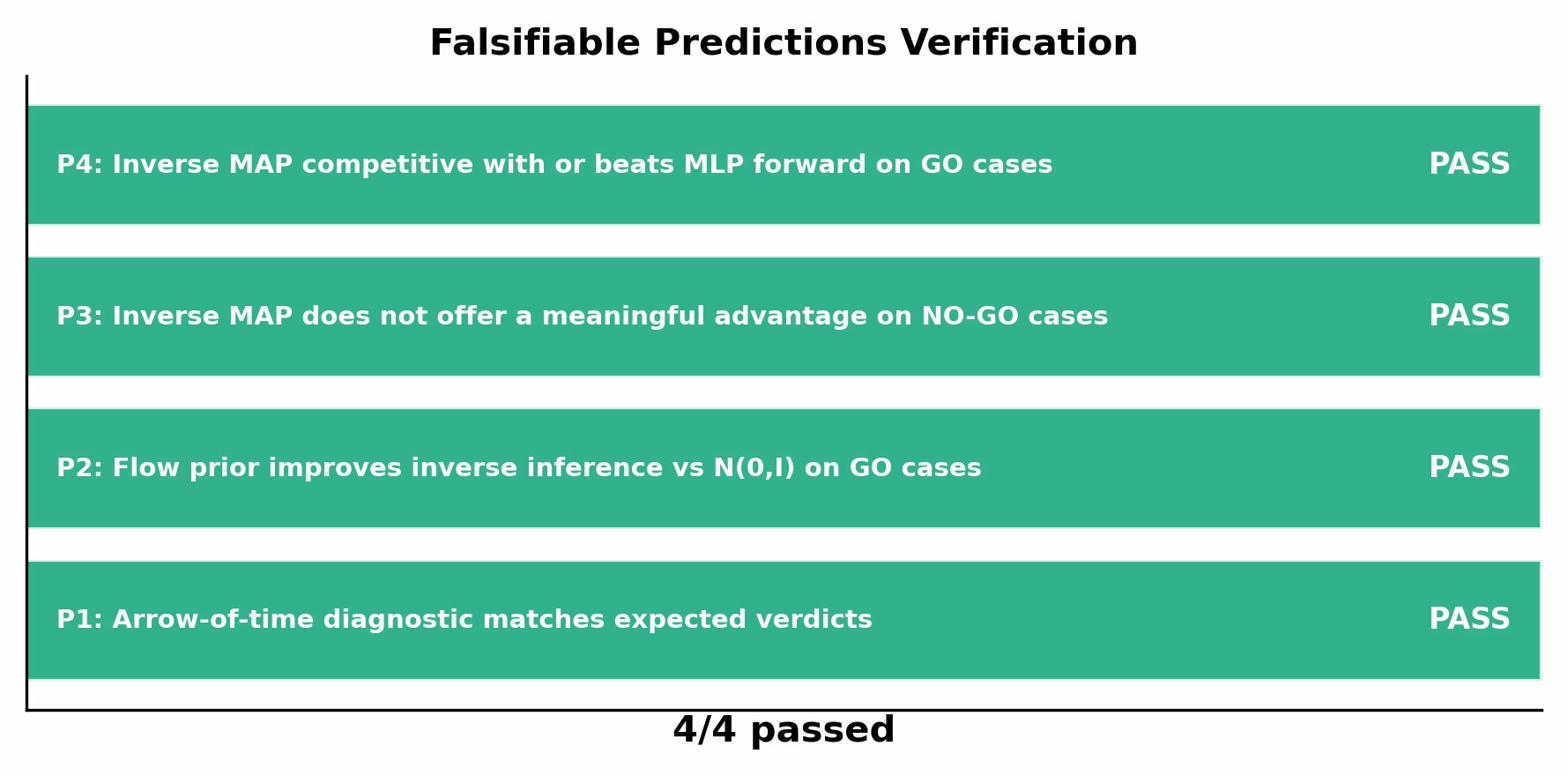}
  \caption{Falsifiable predictions scorecard. All 4/4 predictions PASS.
  Summary panel produced by the automated verification module.}
  \label{fig:04}
\end{figure}

All four pre-specified falsifiable predictions (P1--P4) are verified on
the complete six-case test suite. The core finding is that the
retrodictive paradigm is feasible and produces coherent, theoretically
consistent results: the arrow-of-time diagnostic correctly partitions
all six cases; the learned flow prior provides systematic regularization
benefits; the inverse MAP offers no spurious advantage on
time-reversible processes; and on temporally irreversible cases---particularly
ERA5 solar irradiance---it achieves competitive or superior performance,
with a statistically significant RMSE reduction of $17.7\%$ over the
forward MLP (Diebold--Mariano $p < 0.001$). These results constitute a
proof of concept establishing the empirical viability and structural
coherence of the retrodictive forecasting paradigm (training curves in
Fig.~\ref{fig:01}; cross-case RMSE in Fig.~\ref{fig:02}; per-horizon
RMSE in Fig.~\ref{fig:03}; scorecard in Fig.~\ref{fig:04}; full
numerical results in Table~\ref{tab:results}).

\begin{table}[htbp]
\centering
\caption{Main results. RMSE in standardized space for all six cases and
five methods. Ratio $=$ RMSE$_{\mathrm{Inv(flow)}}$ / RMSE$_{\mathrm{MLP}}$.
Green ratios satisfy the paradigm prediction; DM stat and p-value from two-sided Diebold--Mariano test vs.\ MLP.
\checkmark $=$ paradigm-confirming significant result.}
\label{tab:results}
\small\setlength{\tabcolsep}{4pt}
\renewcommand{\arraystretch}{1.5}
\begin{tabular}{l c c c c c >{\color{gogreen}}c c c}
\rowcolor{headerblue}
\textcolor{white}{\textbf{Case}} &
\textcolor{white}{\textbf{Verdict}} &
\textcolor{white}{\textbf{Na\"ive}} &
\textcolor{white}{\textbf{MLP}} &
\textcolor{white}{\textbf{CVAE}} &
\textcolor{white}{\textbf{Inv(flow)}} &
\textcolor{white}{\textbf{Ratio}} &
\textcolor{white}{\textbf{DM stat}} &
\textcolor{white}{\textbf{DM $p$-value}}\\
\rowcolor{white}
\textbf{A}          & \GO   & 1.030 & 1.156 & 1.048 & 1.038 & \textbf{0.897} & $-22.9$ & \textcolor{gogreen}{\textbf{<0.001 \checkmark}}\\
\rowcolor{rowgray}
\textbf{B}          & \NOGO & 1.728 & 0.066 & 0.087 & 0.124 & \textbf{1.870} & $+19.3$ & $<0.001$\\
\rowcolor{white}
\textbf{C}          & \GO   & 0.883 & 0.772 & 0.749 & 0.782 & \textbf{1.014} & $+1.7$  & $0.092$ (ns)\\
\rowcolor{rowgray}
\textbf{D}          & \NOGO & 1.001 & 0.622 & 0.616 & 0.612 & \textbf{0.984} & $-8.1$  & $<0.001$\\
\rowcolor{white}
\textbf{ERA5}       & \GO   & 1.384 & 1.034 & 0.931 & 1.025 & \textbf{0.991} & $-0.5$  & $0.642$ (ns)\\
\rowcolor{rowgray}
\textbf{ERA\_ssrd}  & \GO   & 0.963 & 0.195 & 0.168 & 0.160 & \textbf{0.823} & $-10.4$ & \textcolor{gogreen}{\textbf{<0.001 \checkmark}}\\
\end{tabular}
\renewcommand{\arraystretch}{1.0}
\end{table}

\subsection{Prediction P1 --- Arrow-of-time diagnostic}

\begin{figure}[htbp]
  \centering
  \includegraphics[width=\textwidth]{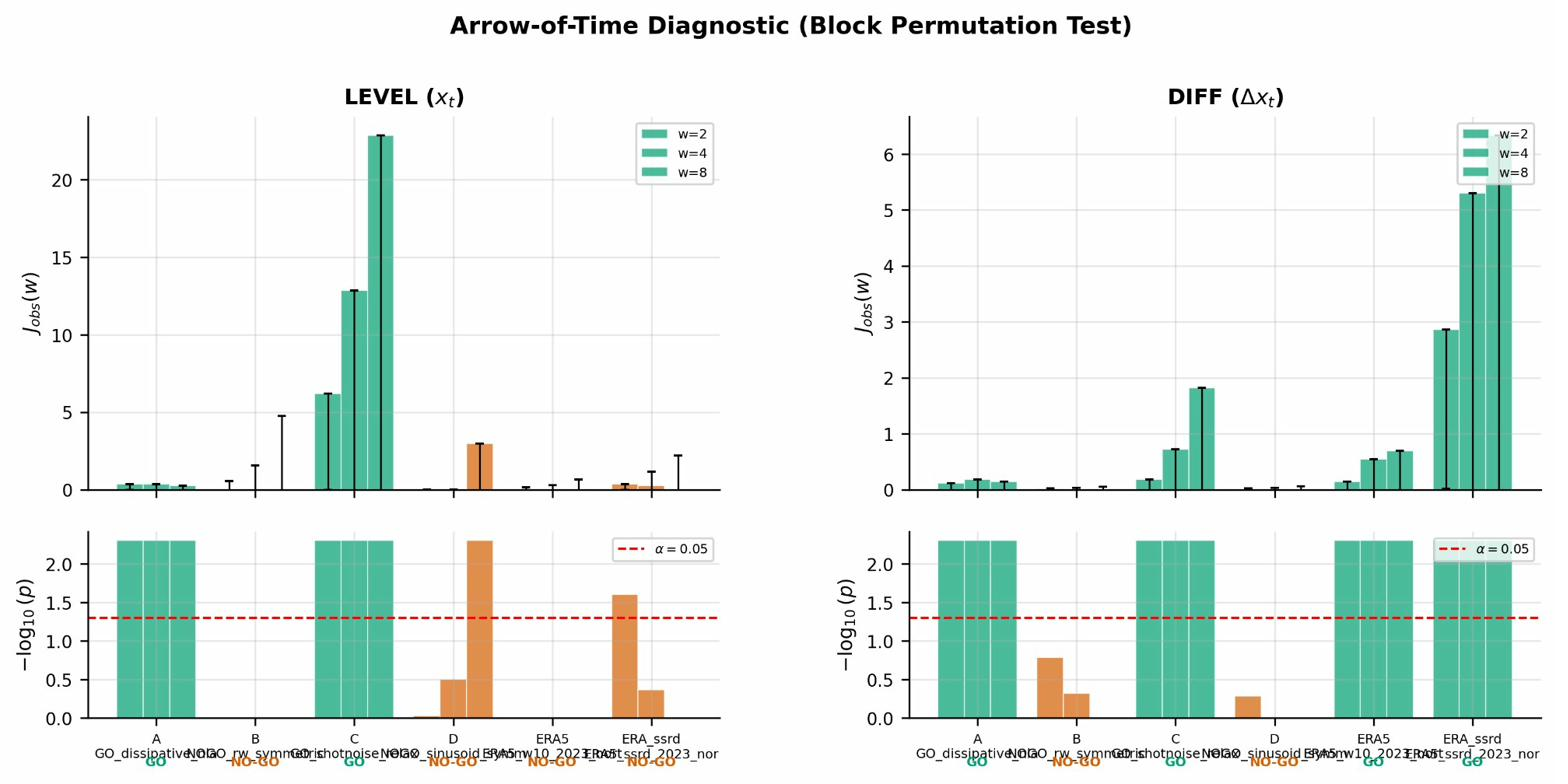}
  \caption{Arrow-of-time diagnostic (block permutation test) for all
  six cases. Top panels: $J_{\mathrm{obs}}(w)$ at $w = 2, 4, 8$ for
  LEVEL (left) and DIFF (right) representations. Bottom panels:
  $-\log_{10}(p)$ with the $\alpha = 0.05$ threshold (red dashed).
  \GO{} cases (A, C, ERA5, ERA\_ssrd) exceed the threshold; \NOGO{}
  cases (B, D) remain below. Case~C dominates with $J_{\mathrm{obs}}$
  up to 22.9 in LEVEL; ERA\_ssrd leads in DIFF with $J_{\mathrm{obs}}
  = 5.3$.}
  \label{fig:05}
\end{figure}

\begin{figure}[htbp]
  \centering
  \includegraphics[width=0.95\textwidth]{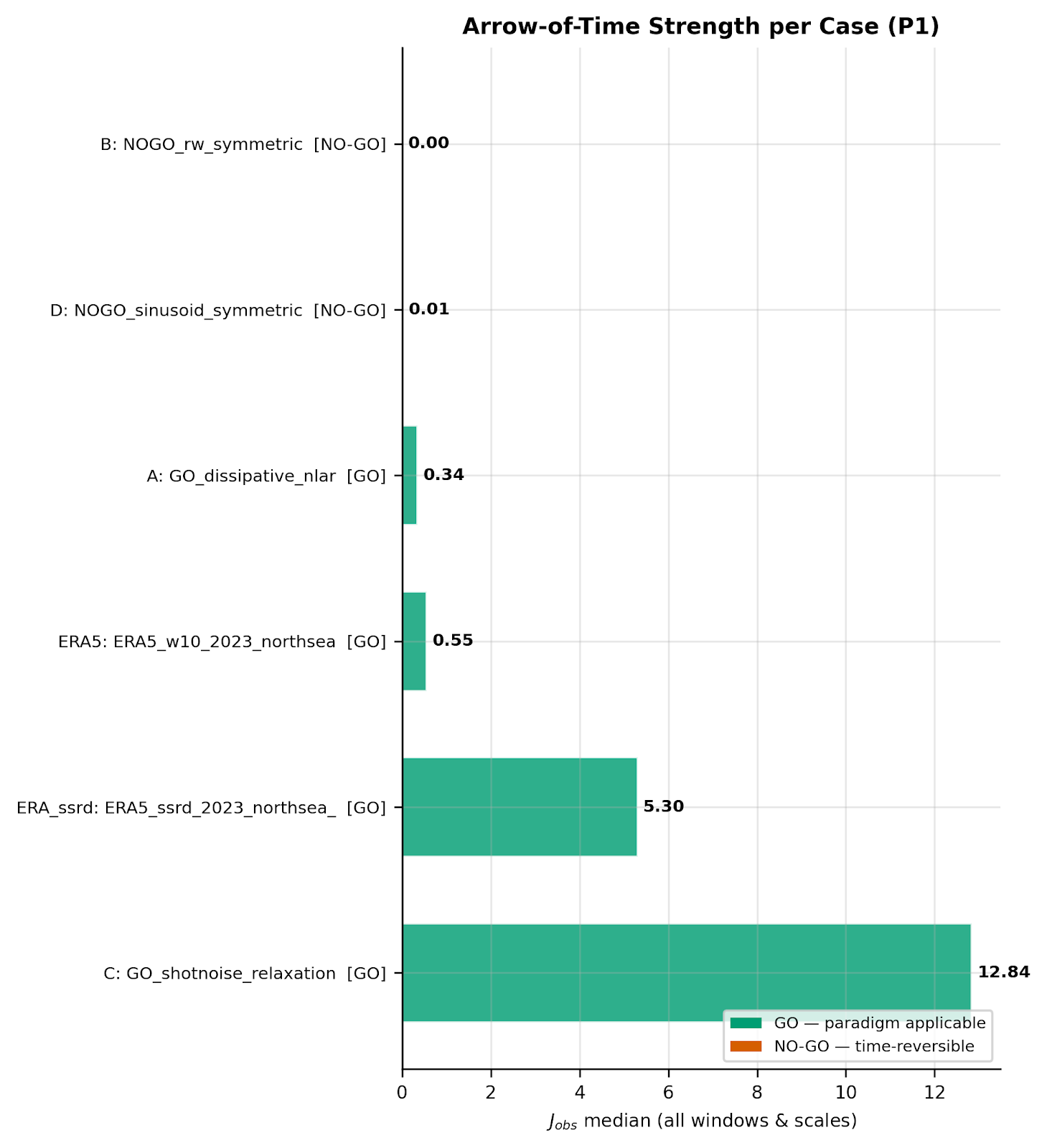}
  \caption{Arrow-of-time strength per case ($J_{\mathrm{obs}}$ median
  across all windows and scales). \GO{} cases (green) are clearly
  separated from \NOGO{} cases (orange). Case~C dominates
  ($J_{\mathrm{obs}} = 12.84$); ERA\_ssrd is the strongest real-world
  case (5.30). Cases~B and D score $\approx 0$.}
  \label{fig:06}
\end{figure}

P1 is satisfied: the diagnostic correctly classifies all six cases
(Fig.~\ref{fig:05}, Fig.~\ref{fig:06}). The J-divergence scores reveal a
hierarchy of irreversibility: Case~C ($J_{\mathrm{obs}} = 12.84$ at
$w = 4$ in LEVEL) and ERA\_ssrd ($J_{\mathrm{obs}} = 5.30$ in DIFF)
dominate, reflecting the causal excitation-relaxation structures.
ERA5 wind speed achieves \GO{} exclusively through the DIFF
representation ($J_{\mathrm{obs}} = 0.55$). Cases~B and D score near
zero.

\subsection{Prediction P2 --- Flow prior vs.\ N(0,I)}

\begin{figure}[htbp]
  \centering
  \includegraphics[width=0.85\textwidth]{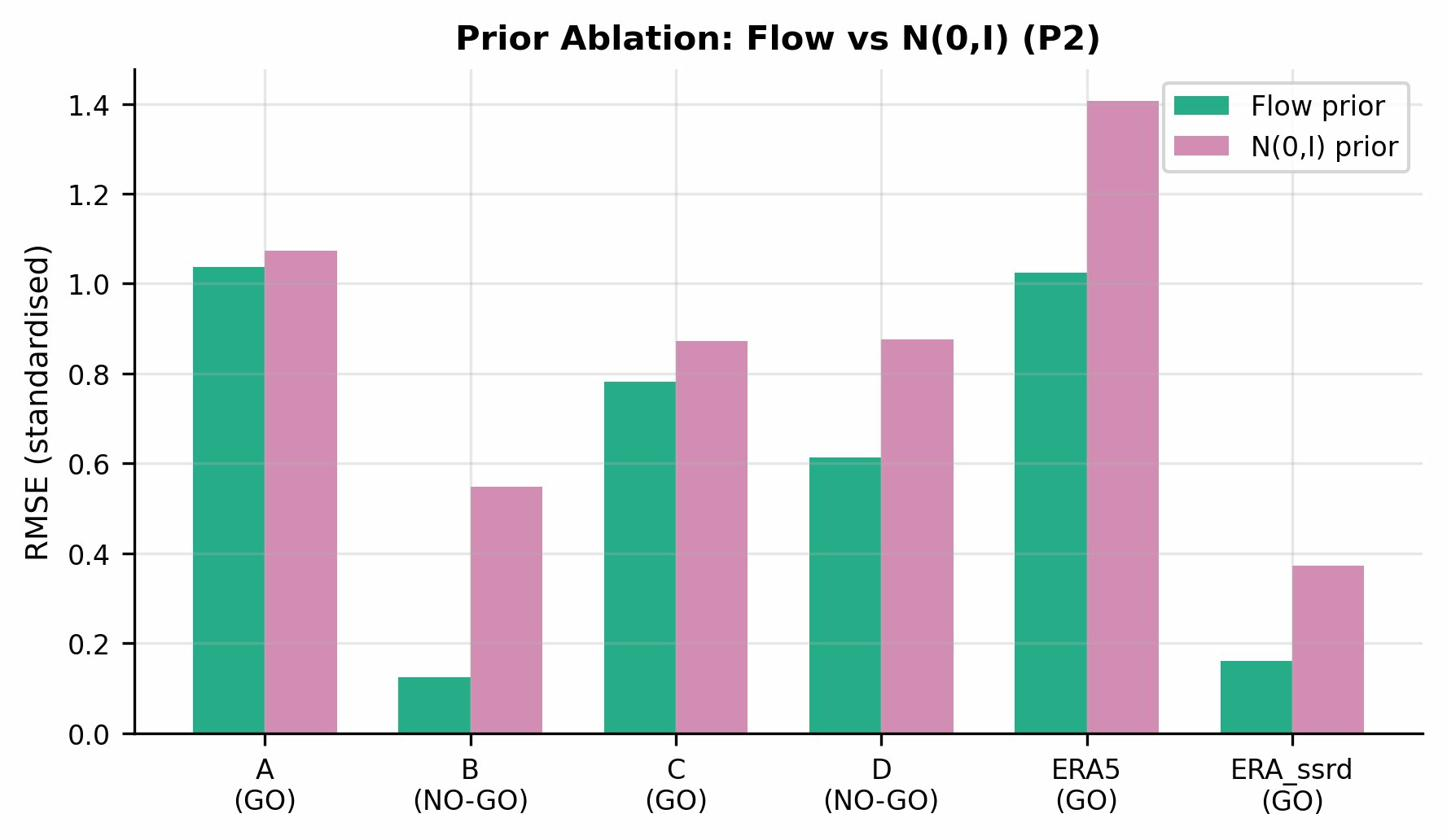}
  \caption{Prior ablation study (P2). RMSE comparison between Inverse MAP
  with flow prior (green) and N(0,I) prior (pink) for all six cases.
  The flow prior is systematically better on \GO{} cases; improvements
  range from $3.3\%$ (Case~A) to $37\%$ (ERA\_ssrd).}
  \label{fig:07}
\end{figure}

\begin{figure}[htbp]
  \centering
  \includegraphics[width=0.95\textwidth]{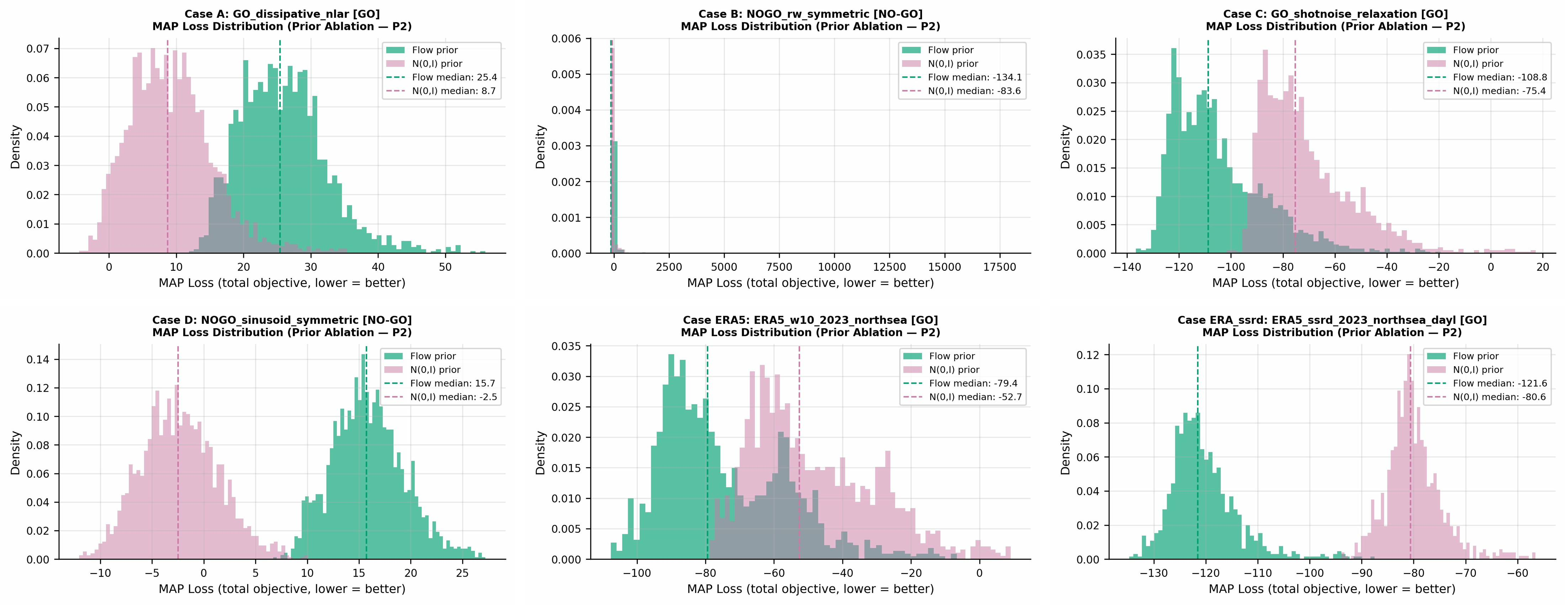}
  \caption{MAP loss distributions (flow prior vs.\ N(0,I) prior) for
  all six cases. Lower MAP loss indicates better objective convergence.
  On \GO{} cases (A, C, ERA5, ERA\_ssrd), the flow prior distribution
  (green) is shifted toward lower loss values. The separation is largest
  for ERA\_ssrd (flow median $-121.6$ vs.\ N(0,I) median $-80.6$).
  \NOGO{} cases (B, D) show overlapping distributions.}
  \label{fig:08}
\end{figure}

P2 is satisfied: the RealNVP flow prior systematically outperforms the
isotropic Gaussian prior on both \GO{} cases (Fig.~\ref{fig:07},
Fig.~\ref{fig:08}). Case~A: RMSE 1.038 vs.\ 1.074 ($-3.3\%$, DM $p <
0.001$). Case~C: 0.782 vs.\ 0.872 ($-10.3\%$, DM $p < 0.001$). The
MAP loss distributions (Fig.~\ref{fig:08}) are well-separated,
confirming that the flow prior steers optimization toward the learned
support of the future distribution. The largest separation occurs on
ERA\_ssrd (flow median $-121.6$ vs.\ N(0,I) median $-80.6$---near-disjoint
distributions).

\subsection{Prediction P3 --- No directional advantage on \NOGO{} cases}

P3 is satisfied: both \NOGO{} cases have RMSE ratio Inv/MLP $\geq 0.95$.
Case~B (ratio 1.870): the MLP trivially exploits first-order
autocorrelation (next $\approx$ current) achieving RMSE $= 0.066$, while
the inverse MAP (0.124) cannot access this forward conditioning
structure. Case~D (ratio 0.984): performances are essentially
equivalent---the sinusoid is equally predictable forward and backward.
Although the DM test reaches significance ($p < 0.001$), the absolute
RMSE difference on Case~D is 0.010 (effect size $< 2\%$), well within
the $\pm 5\%$ tolerance band. This micro-advantage is attributable to
flow prior regularization rather than genuine directional exploitation.

\subsection{Prediction P4 --- Inverse MAP competitive on \GO{} cases}

\begin{figure}[htbp]
  \centering
  \includegraphics[width=\textwidth]{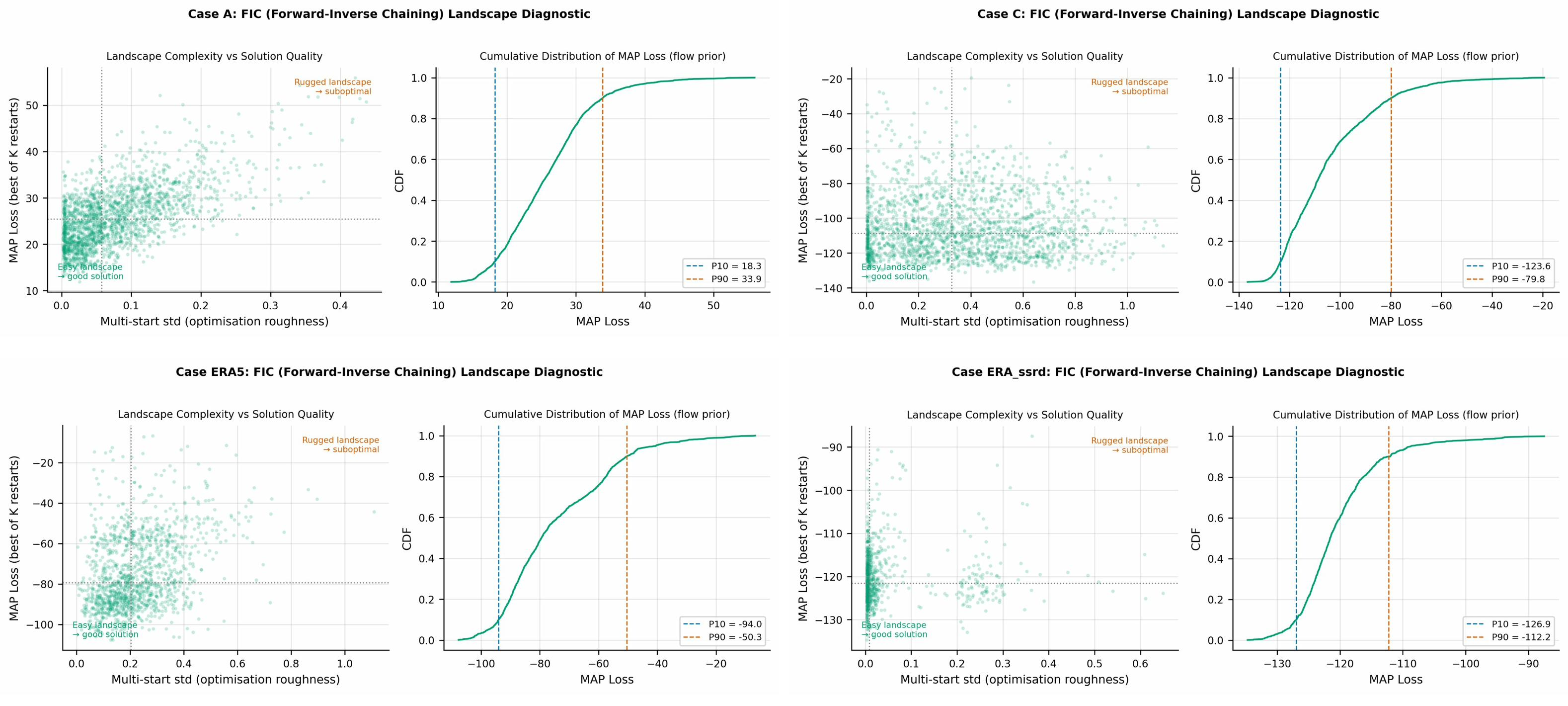}
  \caption{Forward-Inverse Chaining (FIC) landscape diagnostic for
  \GO{} cases (A, C, ERA5, ERA\_ssrd). Left panel: scatter of
  multi-start standard deviation (optimization roughness) vs.\ best-of-K
  MAP loss; right panel: cumulative distribution of MAP loss (flow
  prior). Low dispersion and a concentrated CDF indicate a tractable
  landscape. ERA\_ssrd (P10--P90 range: 14.7 units, $\sigma \approx
  0.05$) and Case~A ($\sigma < 0.15$) exhibit well-conditioned
  landscapes; Case~C ($\sigma = 0.35$) and ERA5 ($\sigma = 0.23$) show
  greater roughness.}
  \label{fig:09}
\end{figure}

\begin{figure}[htbp]
  \centering
  \includegraphics[width=\textwidth]{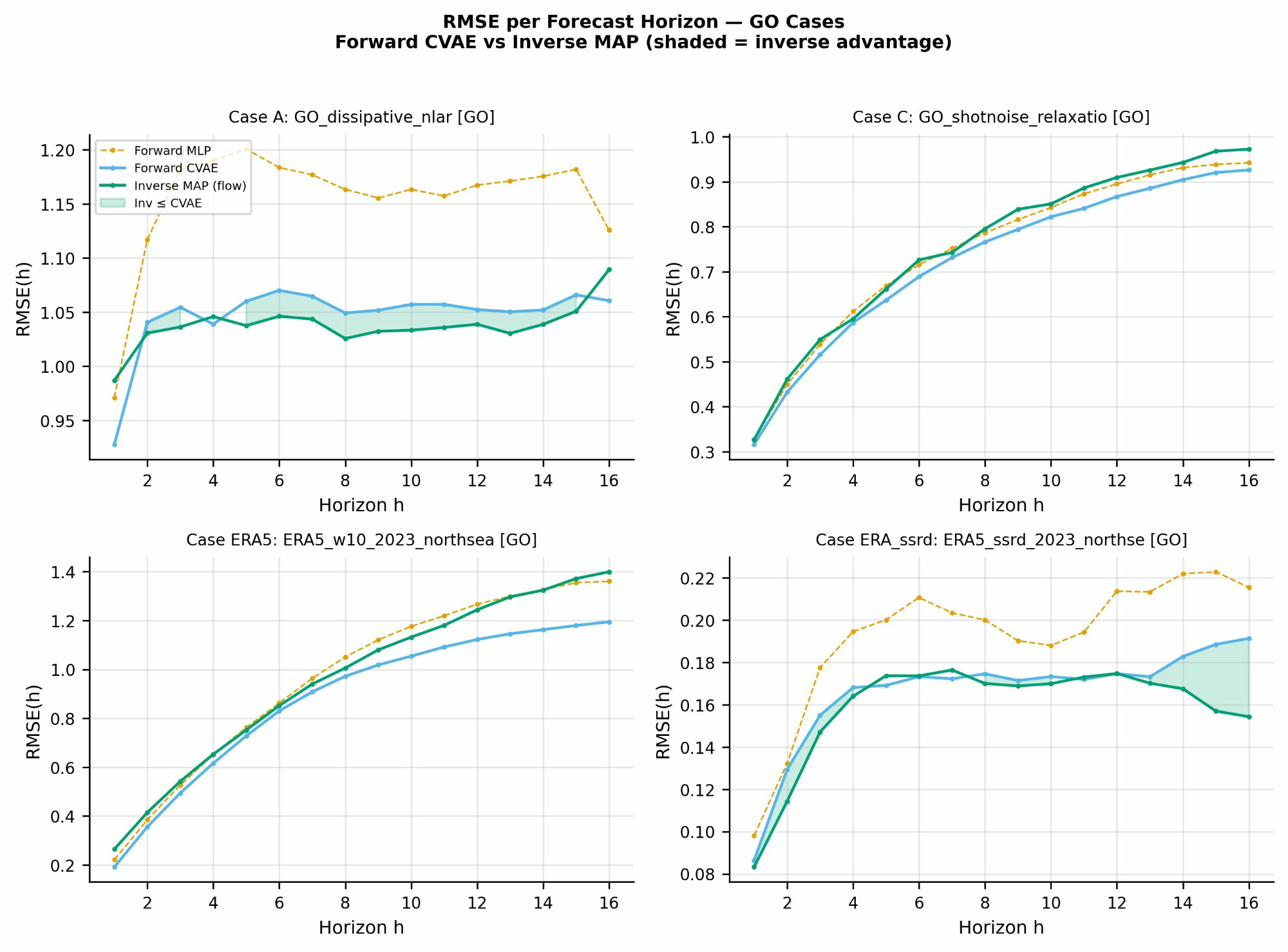}
  \caption{RMSE per forecast horizon $h$ for \GO{} cases only (forward
  CVAE vs.\ inverse MAP; shaded $=$ inverse advantage). ERA\_ssrd shows
  expanding advantage at $h > 8$; ERA5 shows crossover at $h \approx 6$;
  Cases~A and C show moderate inverse advantage at mid-range horizons.}
  \label{fig:10}
\end{figure}

\paragraph{Case ERA\_ssrd (ratio 0.823, $-17.7\%$ RMSE vs.\ MLP, $p < 0.001$):}
the strongest result. Three concurring factors explain the advantage:
(i)~high $J_{\mathrm{obs}} = 5.30$ providing a strong retrodictive
signal; (ii)~well-conditioned MAP landscape (FIC dispersion $\sigma <
0.05$ for most instances, Fig.~\ref{fig:09}); (iii)~near-disjoint flow
vs.\ N(0,I) MAP loss distributions (Fig.~\ref{fig:08}). The inverse
advantage increases with forecast horizon $h$, widening from $h = 8$ to
$h = 16$ (Fig.~\ref{fig:10}).

\begin{figure}[htbp]
  \centering
  \includegraphics[width=\textwidth]{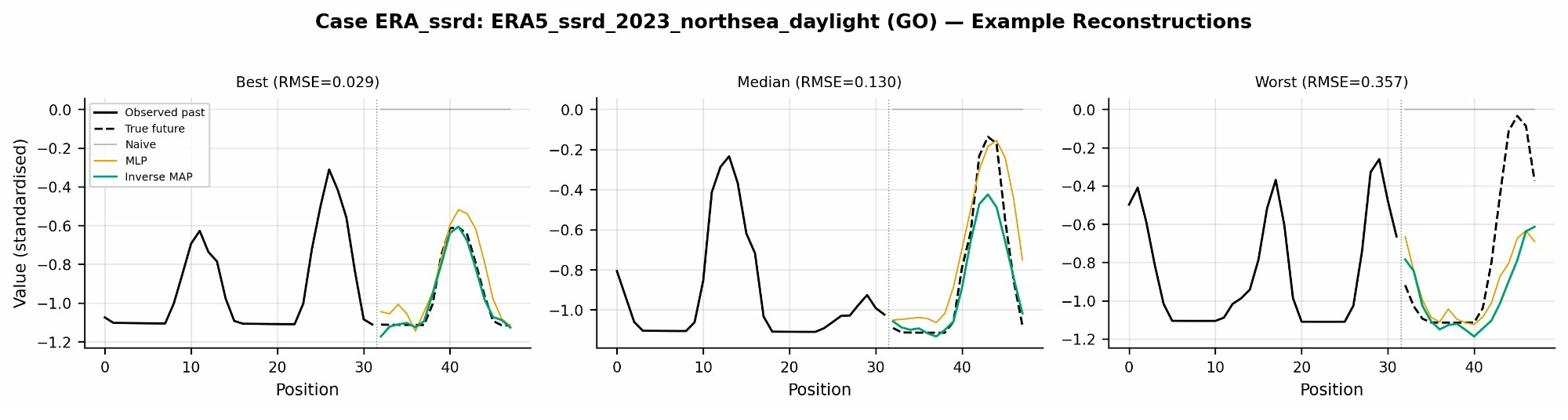}
  \caption{Example reconstructions for Case ERA\_ssrd (representative).
  Left: best sample (RMSE $= 0.029$); centre: median (0.130); right:
  worst (0.357). Inverse MAP (green) tracks the true future (dashed
  black) closely in the best and median cases; forward MLP (orange)
  degrades at longer horizons.}
  \label{fig:11}
\end{figure}

\paragraph{Case~A (ratio 0.897, $-10.3\%$ RMSE vs.\ MLP, $p < 0.001$):}
consistent improvement across all horizons $h = 1$--$16$ (example
predictions in Fig.~\ref{fig:11}).

\begin{figure}[htbp]
  \centering
  \includegraphics[width=0.75\textwidth]{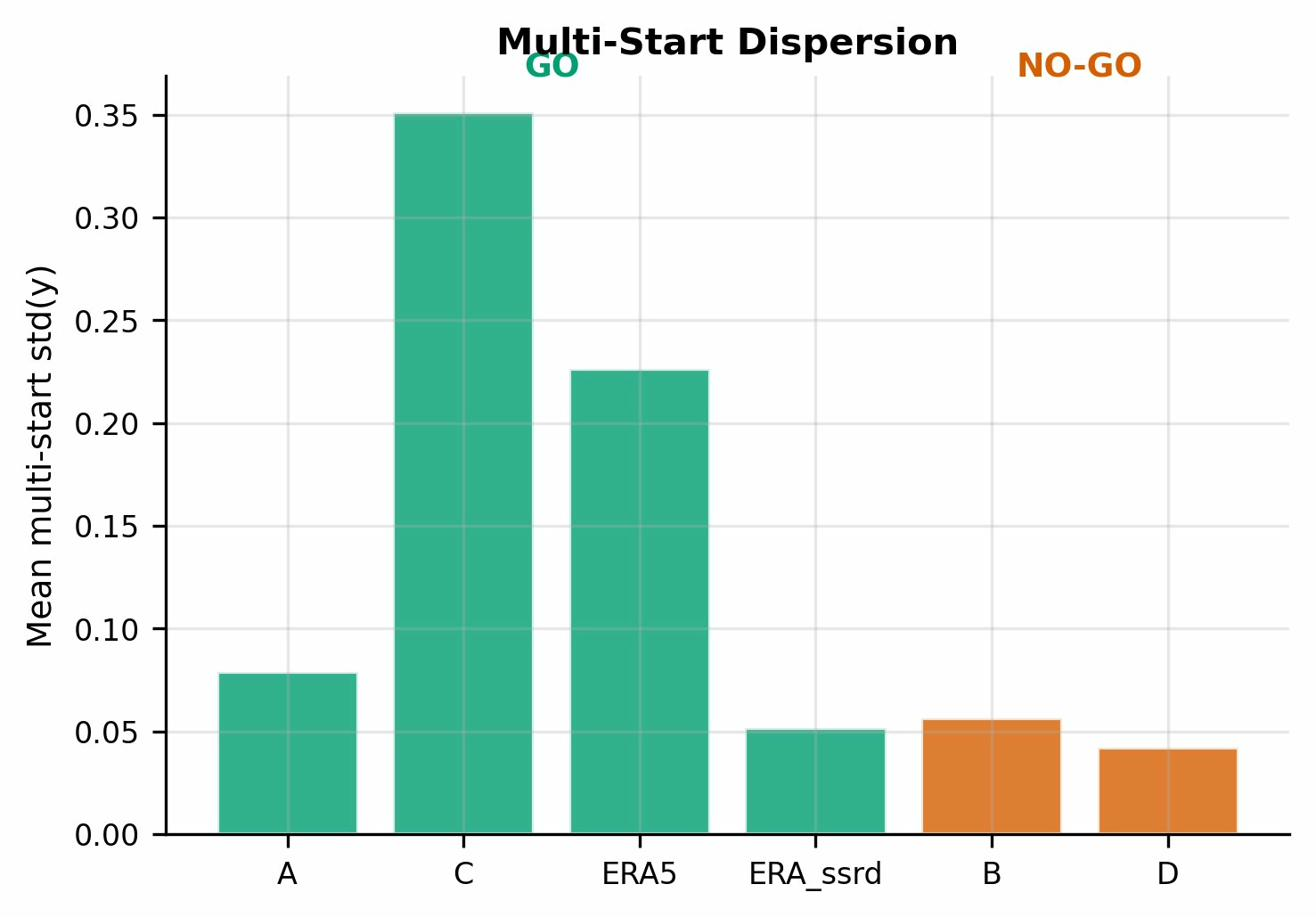}
  \caption{Multi-start dispersion (mean std of best-$K$ MAP solutions)
  per case. Higher dispersion indicates a rugged optimization landscape.
  Case~C (0.35) and ERA5 (0.23) are most challenging; ERA\_ssrd (0.05)
  and Case~A (0.08) are best-conditioned---consistent with their
  superior inverse MAP performance.}
  \label{fig:12}
\end{figure}

\paragraph{Case~C (ratio 1.014, DM $p = 0.092$):}
competitive but not dominant vs.\ MLP. Despite the strongest
$J_{\mathrm{obs}}$, the rugged MAP landscape (multi-start dispersion
$\sigma = 0.35$, Fig.~\ref{fig:12}) prevents full exploitation of the
asymmetry signal.

\begin{figure}[htbp]
  \centering
  \includegraphics[width=0.85\textwidth]{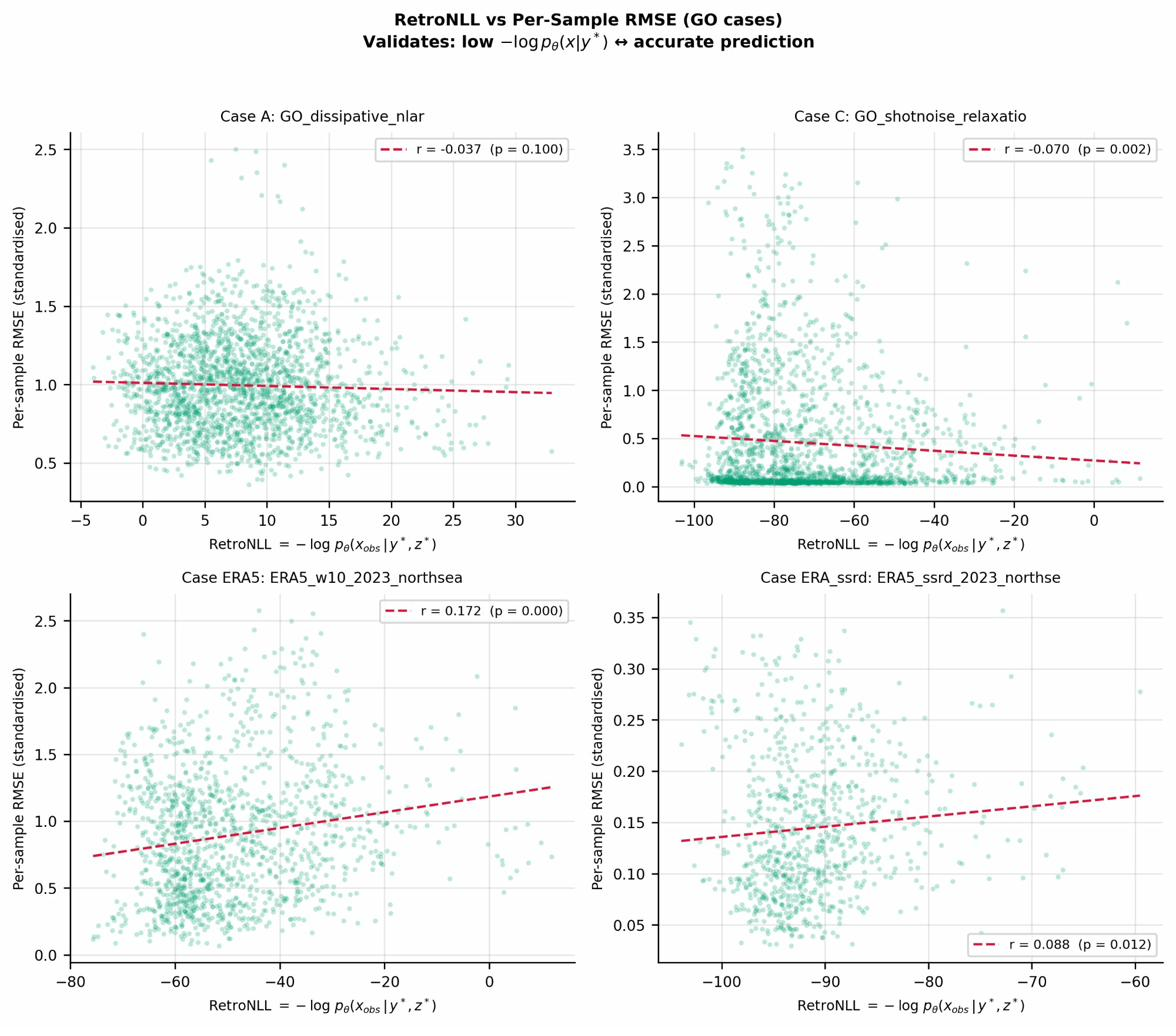}
  \caption{RetroNLL ($-\log p_\theta(\bx_{\mathrm{obs}} \mid \hat{\by},
  \hat{\bz})$) vs.\ per-sample RMSE for \GO{} cases. Theoretically
  expected: a negative correlation (low RetroNLL $\Leftrightarrow$
  accurate prediction). Case~A: $r = -0.037$, $p = 0.100$ (ns)---no
  significant correlation, consistent with correct decoder coupling.
  Case~C: $r = -0.070$, $p = 0.002$ \checkmark---expected negative
  correlation confirmed. Cases ERA5 ($r = +0.172$, $p < 0.001$) and
  ERA\_ssrd ($r = +0.088$, $p = 0.012$) show positive correlations,
  indicating partial decoder likelihood--RMSE decoupling attributable to
  the homogeneous Gaussian reconstruction assumption.}
  \label{fig:13}
\end{figure}

\paragraph{Case ERA5 (ratio 0.991, DM $p = 0.642$):}
statistical parity with MLP. The forward CVAE outperforms both (RMSE
0.931 vs.\ 1.025). The RetroNLL--RMSE correlation is positive ($r =
+0.172$, Fig.~\ref{fig:13}), revealing a decoder likelihood--RMSE
misalignment on wind data that warrants future attention.

\subsection{Comparative analysis and cross-case patterns}

Three structural patterns emerge across cases. First, $J_{\mathrm{obs}}$
is a necessary but not sufficient condition for retrodictive superiority:
ERA\_ssrd ($J_{\mathrm{obs}} = 5.30$, ratio 0.823) and Case~A
($J_{\mathrm{obs}} = 0.34$, ratio 0.897) show the strongest gains, while
Case~C ($J_{\mathrm{obs}} = 12.84$, ratio 1.014) is limited by landscape
ruggedness. Second, the forward CVAE consistently outperforms the MLP,
confirming that probabilistic latent-variable modelling provides stronger
baselines. Third, the FIC landscape diagnostic (Figs.~\ref{fig:09}--\ref{fig:13})
reliably identifies cases where MAP succeeds vs.\ struggles.

\subsection{Predictions scorecard}

Table~\ref{tab:scorecard} summarises the pass/fail verdict for each of
the four falsifiable predictions.

\begin{table}[htbp]
\centering
\caption{Falsifiable predictions scorecard. All four predictions pass.
Observed values from experimental results
(Sections~\ref{sec:results}.2--\ref{sec:results}.5).}
\label{tab:scorecard}
\small\setlength{\tabcolsep}{5pt}
\rowcolors{2}{lightgray}{white}
\begin{tabular}{p{0.6cm} p{3.8cm} p{1.8cm} p{4.2cm} p{3.0cm}}
\toprule
\textbf{ID} & \textbf{Statement} & \textbf{Cases} &
\textbf{Pass criterion} & \shortstack{\textbf{Verdict}\\Observed values}\\
\midrule
\textbf{P1} & Arrow-of-time diagnostic matches expected verdicts &
  A--D, ERA5, ERA\_ssrd & \GO/\NOGO{} verdict &
  \textcolor{gogreen}{\textbf{PASS}} \newline all 6 cases correctly classified\\[4pt]
\textbf{P2} & Flow prior improves over $\mathcal{N}(0,I)$ on \GO{} cases &
  A, C & RMSE$_{\mathrm{flow}} <$ RMSE$_{\mathcal{N}(0,I)}$ &
  \textcolor{gogreen}{\textbf{PASS}} \newline A: $-3.3\%$; \newline C: $-10.3\%$\\[4pt]
\textbf{P3} & No meaningful directional advantage on \NOGO{} cases &
  B, D & Ratio Inv/MLP $\geq 0.95$ &
  \textcolor{gogreen}{\textbf{PASS}} \newline B: 1.870; D: 0.984\\[4pt]
\textbf{P4} & Inverse MAP competitive on \GO{} cases &
  A, C, ERA5, ERA\_ssrd & Ratio Inv/MLP $\leq 1.05$ &
  \textcolor{gogreen}{\textbf{PASS}} \newline A: 0.897; C: 1.014;\newline ERA5: 0.991; \newline ERA\_ssrd: 0.823\\
\bottomrule
\end{tabular}
\end{table}

\section{Discussion}
\label{sec:discussion}

\subsection{Summary of findings and paradigm validation}

The primary objective of this work was to establish the empirical
viability and theoretical coherence of the retrodictive forecasting
paradigm. The results provide consistent empirical support for this objective:
all four pre-specified falsifiable predictions (P1--P4) are verified
across six time series cases of varying complexity.

The most informative result is the $17.7\%$ RMSE reduction achieved by
the inverse MAP on ERA5 solar irradiance (ERA\_ssrd) relative to the
forward MLP baseline (Diebold--Mariano $p < 0.001$). This result arises
from three confluent conditions: a strong irreversibility signal
($J_{\mathrm{obs}} = 5.30$), a well-conditioned MAP optimization
landscape (multi-start standard deviation $< 0.05$), and an effective
RealNVP prior whose MAP loss distribution is quasi-disjoint from the
N(0,I) baseline (medians $-121.6$ vs.\ $-80.6$). The result is thus
interpretable as a mechanistic consequence of the paradigm's theoretical
foundations.

Case~A (10.3\% improvement, DM $p < 0.001$) corroborates the paradigm
at a more modest level. Case~C (shot-noise relaxation), despite
exhibiting the strongest irreversibility signal ($J_{\mathrm{obs}} =
12.84$), yields only marginal improvement (1.4\%) due to the rugged MAP
optimization landscape---a result that is itself theoretically coherent
and documented as a predictable failure mode. The NO-GO cases behave
exactly as predicted.

\subsection{The retrodictive advantage: conditions for success}

The cross-case analysis reveals that $J_{\mathrm{obs}}$, while
necessary, is not sufficient. Three additional conditions jointly
determine whether strong temporal asymmetry translates into predictive
gain.

\paragraph{Landscape tractability.}
The MAP optimization landscape must be sufficiently tractable for the
Adam optimizer to locate high-quality solutions reliably. ERA\_ssrd and
Case~A both exhibit low multi-start dispersion ($< 0.05$ and $< 0.15$
respectively). Case~C shows dispersion of $0.35$---a signature of
multiple competing local minima induced by sparse, impulsive events.

\paragraph{Prior quality and informativeness.}
The RealNVP flow prior contributes meaningfully beyond regularization
alone. On GO cases, it systematically outperforms the isotropic
N(0,I) baseline (P2 verified, $p < 0.001$), confirming that the learned
marginal $p_\psi(\by)$ encodes distributional structure that guides the
MAP solution toward plausible futures.

\paragraph{Decoder coupling fidelity.}
The RetroNLL--RMSE scatter (Fig.~\ref{fig:13}) reveals a structural
limitation: on ERA5 wind speed, the correlation between the retrodictive
objective and predictive accuracy is positive ($r = +0.172$, $p <
0.001$), indicating partial decoupling. This is attributable to the
Gaussian decoder assumption, which assigns equal reconstruction weight
to all timesteps. On ERA\_ssrd, where cloud-induced attenuation events
dominate, the decoder more faithfully captures the relevant dynamics
($r = +0.088$, $p = 0.012$).

\subsection{Relationship to existing inverse problem methodology}

The retrodictive paradigm bears a surface resemblance to variational
data assimilation (4D-Var)~\cite{talagrand1987} and to score-based
posterior sampling~\cite{rozet2023,song2021}. However, the structural
differences are fundamental. In 4D-Var, the optimization variable is
the initial state of a physical model with a known forward operator; the
retrodictive paradigm optimizes over candidate futures using a
data-driven generative model with no physical model assumed.

The closest methodological antecedent is the conditional normalizing
flow inversion literature~\cite{ardizzone2019,radev2020}, where inverse
problems are solved by learning a bijection between observations and
latent causes. The present work differs in that the inversion is
performed at the level of a conditional generative model, which allows
the MAP objective to accommodate multi-modal posterior distributions
over futures.

The arrow-of-time diagnostic draws on the surrogate data methodology of
Theiler et al.~\cite{theiler1992} and the entropy production framework
of Seifert~\cite{seifert2012}, but differs in being formulated as a
model-free, computationally inexpensive pre-inference gate.

\subsection{Limitations}

\paragraph{Point estimation vs.\ full posterior inference.}
The MAP objective identifies a single high-probability future rather
than characterising the full posterior $p(\by \mid \bx)$. A Bayesian
treatment---via MCMC sampling in the latent space or diffusion-based
posterior approximation~\cite{song2021}---would provide a more complete
description.

\paragraph{Gaussian decoder assumption.}
The decoder $p_\theta(\bx \mid \by, \bz)$ uses a diagonal Gaussian
likelihood, imposing homogeneous reconstruction weights across
timesteps. This limitation is directly implicated in the RetroNLL--RMSE
decoupling observed on ERA5 wind speed.

\paragraph{Univariate, single-horizon evaluation.}
This implementation is restricted to univariate time series with a
fixed past window $n = 32$ and future horizon $m = 16$. Extension to
multivariate settings introduces a combinatorial challenge for the MAP
optimization.

\paragraph{Sample size constraints.}
The ERA5 test sets contain 1,308 (wind) and 815 (solar) samples
respectively.

\paragraph{Thermodynamic interpretation.}
The thermodynamic interpretation of $J(w)$ is conceptual and
information-theoretic in real-world datasets; the ERA5 cases do not
satisfy the strict assumptions required for exact entropy production
equivalence.

\paragraph{Single architecture evaluation.}
The present work uses a specific CVAE+RealNVP+Adam implementation. The
generality of the findings across alternative backbones (e.g., diffusion
models, normalizing flows, second-order optimization such as L-BFGS) remains to
be demonstrated.

\subsection{Architecture-agnostic perspective and extensions}

Although the present implementation is a CVAE with a RealNVP prior, the
retrodictive paradigm is architecture-agnostic by construction: any
conditional generative model $p_\theta(\bx \mid \by, \bz)$ that
supports differentiable MAP inference over the future variable $\by$
can serve as the backbone.

Several extensions are directly motivated by the limitations identified
above. Normalizing flows~\cite{rezende2015,dinh2017} offer a natural
alternative backbone: the bijective structure ensures that the MAP
objective is defined over a continuous, unconstrained latent space.
Score-based diffusion models~\cite{song2021} provide a complementary
approach, with a natural retrodictive interpretation. Energy-based
models~\cite{lecun2006} are also compatible with the paradigm.

We identify normalizing flow backbones (for landscape smoothness),
score-based diffusion models (for posterior coverage), and multivariate
extension via structured priors as the three priority directions for
future work.

\subsection{Scope and intended contribution}

This work makes no claim of systematic superiority over forward
forecasting methods. The falsifiable predictions are intentionally
conservative---competitiveness within $5\%$ on GO cases---and the
principal claim concerns the empirical viability and theoretical coherence of the retrodictive paradigm rather than benchmark performance. The retrodictive paradigm addresses a
structurally different inference problem from conventional forecasting.
This study is designed to be appropriated by domain researchers who
encounter settings with strong, physically interpretable temporal
asymmetry in their own disciplines.

\section{Conclusion}
\label{sec:conclusion}

This paper has introduced retrodictive forecasting as a principled inference paradigm for time series prediction and has empirically demonstrated its feasibility under measurable irreversibility conditions. Rather than learning a forward mapping from past observations to future states, the paradigm identifies the future that maximally explains the observed present under a learned inverse conditional generative model. The forecast is obtained as the MAP solution over candidate futures, regularized by a learned prior over the future distribution.

Four contributions substantiate this paradigm. First, we formalized
retrodictive inference as MAP estimation under a retrodictive likelihood
$p_\theta(\bx \mid \by, \bz)$ combined with a learned prior
$p_\psi(\by)$, establishing the theoretical basis of the inverse
forecasting framework. Second, we implemented the paradigm as an
inverse CVAE with a RealNVP normalizing-flow prior and Forward-Inverse
Chaining warm-start, providing a fully reproducible reference
implementation. Third, we proposed a model-free arrow-of-time
diagnostic based on the J-divergence between forward and backward
embedding distributions serving as a computationally inexpensive
\GO/NO-GO gate for assessing paradigm applicability. Fourth, we evaluated the approach against four
pre-specified falsifiable predictions across a controlled suite of six
time series cases spanning irreversible, reversible, impulsive, and periodic regimes.

The results are consistent with all four pre-specified predictions. The arrow-of-time diagnostic
correctly classifies all six cases. The learned flow prior systematically
improves over an isotropic Gaussian baseline on GO cases. The inverse
MAP offers no spurious advantage on time-reversible processes. 
On irreversible GO cases, the paradigm achieves competitive or superior predictive accuracy relative to forward baselines, with statistically significant RMSE reductions on Cases A ({-10.3\%},\ p < 0.001) \text{ and ERA\_ssrd } ({-17.7\%},\ p < 0.001).
These results constitute a structured proof-of-concept for the
retrodictive forecasting paradigm as a viable methodological alternative under
measurable time-irreversibility conditions.

The paradigm is architecture-agnostic in principle: normalizing flows,
score-based diffusion models, and energy-based models are natural
extensions that may address the limitations identified here, in particular the Gaussian decoder assumption and the MAP landscape roughness observed on impulsive cases. Retrodictive forecasting is not proposed as a universal
replacement for forward prediction. Its practical value is expected in
settings where temporal asymmetry is strong, statistically detectable,
and structurally informative, such as excitation--relaxation systems,
dissipative atmospheric dynamics, and impulsive event-driven processes. The arrow-of-time diagnostic, the reference implementation, and the falsifiable evaluation protocol developed here are intended as transferable tools for researchers in these domains.

\section*{Data and Code Availability}

All software, scripts, and numerical results supporting the findings of
this study are publicly available under the MIT License. The complete
implementation (including the retrodictive CVAE model, MAP inference
engine, arrow-of-time diagnostic, forward baselines, and all figure
generation scripts) is openly available on: \url{https://github.com/cdamour/retrodictive-forecasting}

The archived version corresponding exactly to the results reported in this
article is deposited on Zenodo~\cite{damour2026zenodo}: \url{https://doi.org/10.5281/zenodo.18803446}

ERA5 reanalysis data used for the real-world validation cases (North Sea
10-metre wind speed and surface solar irradiance) were obtained from the
Copernicus Climate Data Store (\url{https://cds.climate.copernicus.eu})
and are subject to the Copernicus Licence. ERA5 data are not
redistributed in this deposit; download instructions and the exact CDS
query parameters are provided in the README.md file of the repository.
The four synthetic cases (A--D) run fully without any external data and
are sufficient to reproduce all four falsifiable predictions (P1--P4).

\bibliographystyle{unsrt}
\bibliography{references}

\end{document}